\documentclass[sigconf]{acmart}

\AtBeginDocument{%
  \providecommand\BibTeX{{%
    \normalfont B\kern-0.5em{\scshape i\kern-0.25em b}\kern-0.8em\TeX}}}

\setcopyright{acmlicensed}
\copyrightyear{2018}
\acmYear{2018}
\acmDOI{XXXXXXX.XXXXXXX}

\acmConference[Conference acronym 'XX]{Make sure to enter the correct
  conference title from your rights confirmation emai}{June 03--05,
  2018}{Woodstock, NY}
\acmISBN{978-1-4503-XXXX-X/18/06}

\usepackage{cases}
\usepackage{multirow}
\usepackage{subfigure}
\begin{document}

\title{Self-Distilled Disentangled Learning for Counterfactual Prediction}

\author{Xinshu Li}
\email{xinshu.li@unsw.edu.au}
\affiliation{%
  \institution{The University of New South Wales}
  \city{Sydney}
  \country{Australia}
}
\author{Mingming Gong}
\email{mingming.gong@unimelb.edu.au}
\affiliation{%
  \institution{The University of Melbourne}
  \city{Melbourne}
  \country{Australia}
}
\affiliation{%
  \institution{MBZUAI}
  \city{Abu Dhabi}
  \country{United Arab Emirates}
}
\author{Lina Yao}
\email{lina.yao@data61.csiro.au}
\affiliation{%
  \institution{CSIRO’s Data 61}
  \institution{The University of New South Wales}
  \city{Sydney}
  \country{Australia}
}
\begin{abstract}
  The advancements in disentangled representation learning significantly enhance the accuracy of counterfactual predictions by granting precise control over instrumental variables, confounders, and adjustable variables. An appealing method for achieving the independent separation of these factors is mutual information minimization, a task that presents challenges in numerous machine learning scenarios, especially within high-dimensional spaces. To circumvent this challenge, we propose the Self-Distilled Disentanglement framework, referred to as $SD^2$. Grounded in information theory, it ensures theoretically sound independent disentangled representations without intricate mutual information estimator designs for high-dimensional representations. Our comprehensive experiments, conducted on both synthetic and real-world datasets, confirms the effectiveness of our approach in facilitating counterfactual inference in the presence of both observed and unobserved confounders.
\end{abstract}






\maketitle

\section{Introduction}
Counterfactual prediction has attracted increasing attention \citep{alaa2017bayesian,  chernozhukov2013inference, glass2013causal,zhou2023emerging} in recent years due to the rising demands for robust and trustworthy artificial intelligence. Confounders, the common causes of treatments and effects, induce spurious relations between different variables, consequently undermining the distillation of causal relations from associations. Thanks to the rapid development of representation learning, a plethora of methods \citep{10027640,yao2018representation, shalit2017estimating} mitigate the bias caused by the observed confounders via generating balanced representations in the latent space. As for the bias brought by the unobserved confounders, \citet{pearl2000models, Angrist1994IdentificationAE,hartford2017deep,muandet2020dual,lin2019one} propose obtaining an unbiased estimator by regressing outcomes on Instrumental variables (IVs), which are exogenous variables related to treatment and only affect outcomes indirectly via treatment, to break the information flow between unobserved confounders and the treatments.  

\begin{figure}[t!]
\centering{\includegraphics[width=\columnwidth]{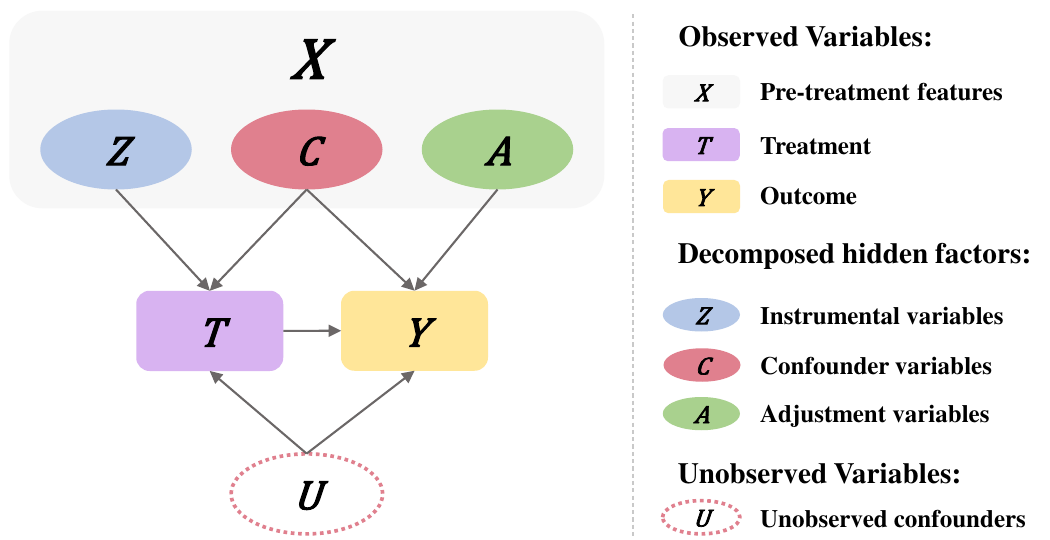}}
\caption{General causal structure. The underlying confounders $C$ in observed pre-treatment features $X$ and unobserved confounders $U$ result in spurious relations rather than causal relations between treatment $T$ and outcome $Y$. We aim to disentangle mutually independent representations of $Z$, $C$, and $A$ from $X$ without the design of intrigue mutual information estimators. }
\label{causal graph}
\end{figure}

We will further clarify by integrating Figure \ref{causal graph} with a real-world example. \citet{hoxby2000does} examined whether competition among public schools (Treatment T) improves educational quality within districts (Outcome Y). The potential endogeneity of the number of schools in a region arises because both the treatment and the outcome could be influenced by long-term factors (Confounders) specific to the area. Some of these confounders (observed Confounders C) can be measured, such as the level of local economic development. However, other confounders (Unobserved Confounders U) cannot be easily quantified or exhaustively listed, such as certain historical factors. River counts here can serve as a persuasive IV: 1) more rivers can lead to more schools due to transportation (Relevance); 2) yet they are unrelated to teaching quality directly (Exclusivity); 3) the natural attributes of river formation render the counts of the rivers unrelated to the confounders caused by historical factors (Exogeneity).

There are two drawbacks to the existing methods. Firstly, most of them treat all observed features as the observed confounders to block, while only part contributes to the distribution discrepancy of pre-treatment features. Secondly, the prerequisite for IV-based regression methods is to access valid IVs, which have three strict conditions (Relevance, Exclusion, Exogeneity) to satisfy, as shown above, making it a thorny task to find.

Disentangled representation learning \citep{hassanpour2019learning,wu2020learning,yuan2022auto,cheng2022learning}, aiming at decomposing the representations of different underlying factors from the observed features, is showing promise in addressing the flaws above simultaneously. The disentangled factors provide us with a more precise inference route to alleviate the bias brought by the observed confounders. Additionally, one can automatically obtain the representations of the valid IVs to remove the bias led by the unobserved confounders. 

The seminar work by \citet{hassanpour2019learning} introduces disentangled learning to the domain of counterfactual prediction, creatively categorizing observed features into instrumental variables (IVs), confounders, and adjustable variables. However, it does not guarantee the generation of mutually independent representations of these underlying factors. Subsequently, \citet{cheng2022learning,wu2020learning,yuan2022auto} integrate effective mutual information estimators \citep{Cheng2020CLUBAC} to minimize the mutual information between latent factor representations, aiming to achieve mutually independent latent representations. Nonetheless, the accurate estimation of mutual information between high-dimensional representations remains a persistent challenge \citep{belghazi2018mutual,poole2019variational}. Inaccurate estimation of mutual information may result in the failure to obtain mutually independent representations of latent factors, thereby compromising the accuracy of counterfactual estimation.

To overcome the defects of previous methods, we propose a novel algorithm for counterfactual prediction named Self-Distilled Disentanglement ($SD^2$) to sidestep MI estimation between high-dimensional representations. Specifically, we provide theoretical analysis rooted in information theory to guarantee the mutual independence of diverse underlying factors. Further, we give a solvable form of our method through rigorous mathematical derivation to minimize MI directly rather than explicitly estimating it. Based on the theory put forward, we design a hierarchical distillation framework to kill three birds with one stone: disentangle three independent underlying factors, mitigate the confounding bias, and grasp sufficient supervision information for counterfactual prediction.

Our main contributions are summarized as follows:

\begin{itemize}
\item We put forward a theoretically assured solution rooted in information theory for disentanglement to avoid mutual information estimation between high-dimensional representations. It enables the generation of mutually independent representations, thereby adding in the decomposition of diverse underlying factors.
\item We provide a tractable form of proposed solution through mathematically rigorous derivation, which rewrites the loss function of disentanglement and fundamentally tackles the difficulty of mutual information estimation between high-dimensional representations.
\item We propose a novel self-distilled disentanglement method ($SD^2$) for counterfactual prediction. By designing a hierarchical distillation framework, we disentangle IVs and confounders from observational data to mitigate the bias induced by the observed and unobserved confounders at the same time.
\item We conduct extensive experiments on synthetic and real-world benchmarks to verify our theoretically grounded strategies. The results demonstrate the effectiveness of our framework on counterfactual prediction compared with the state-of-the-art baselines. 
\end{itemize}

\section{Related Work}
\label{related}

\subsection{Counterfactual Prediction}
The main challenge for counterfactual prediction is the existence of confounders. Researchers adopt matching \citep{rosenbaum1983central}, re-weighting \citep{imbens2004nonparametric}, regression \citep{chipman2010bart}, representation learning \citep{shalit2017estimating,zhou2022cycle} to alleviate the confounding bias under the ``no hidden confounders" assumption. To relax this unpractical assumption, a few non-parametric or semi-parametric methods utilize special structures among the variables to resolve the bias led by the unobserved confounders. These structures include (1) proxy variables \citep{veitch2019using,veitch2020adapting,Louizos2017CausalEI,wu2022betaintactvae}; (2) multiple causes \citep{wang2019blessings,zhang2019medical,cheng2022effects}; (3) instrumental variables. 2SLS \citep{Angrist1994IdentificationAE} is the classical IV method in a linear setting. \citet{singh2019kernel,wu2022instrumental,xu2020learning,hartford2017deep,lin2019one} adopt advanced machine learning or deep learning algorithms for non-linear scenarios. Another commonly used causal effects estimator using IVs is the control function estimator (CFN) \citep{Wooldridge420,puli2020general}. These methods require well-predefined IVs or assume all observed features as confounders, which inevitably impairs their generalization to real-world practice. \citet{yuan2022auto,Wu2022TreatmentEE,li2024distribution} aims to generate IVs for downstream IV-based methods rather than focusing on counterfactual prediction.

\subsection{Disentangled Representation Learning}
Most of the current state-of-the-art disentangled representation learning methods are based on \citet{Kingma2013AutoEncodingVB}, which uses a variational approximation posterior for the inference process of the latent variables. Typical work includes $\beta$-VAE \citep{higgins2017betavae}, Factor-VAE \citep{Kim2018DisentanglingBF}, CEVAE \citep{Louizos2017CausalEI} and so on. Another popular solution \citep{Chen2016InfoGANIR,Zou2020JointDA} builds on the basis of Generative Adversarial Networks \citep{Goodfellow2014GenerativeAN}. However, these studies are more suitable for the approximate data generation problem and fall short when it comes to estimating causal effects, largely due to the difficulty in training complex generation models. 

As the disentanglement of the underlying factors in observed features helps to alleviate confounding bias and improve inference accuracy, \citet{hassanpour2019learning,zhang2021treatment,wu2020learning} design different decomposition regularizers to help the separation of underlying factors. However, these methods fall short in obtaining disentangled representations that are mutually independent, which is a prerequisite for identifying treatment effects. 
\citet{yuan2022auto,cheng2022learning} employ a cutting-edge mutual information estimator \citep{Cheng2020CLUBAC} to reduce the mutual information between disentangled representations, thereby promoting independence among the representations of underlying factors. Despite this, the intricacy of mutual information estimation poses a challenge to the accurate separation of latent factors, thereby leaving room for enhancement in counterfactual prediction. 

 Compared with previous IV-based counterfactual prediction methods, $SD^2$ abandons well-predefined IVs but rather decomposes them from pre-treatment variables. Besides, we only impose conditions on variables that are related to confounding bias. Hence, our approach effectively mitigates both unobserved and observed confounding biases concurrently, a feat rarely achieved by previous IV-based counterfactual prediction methods. Compared with generative disentangled methods, rather than generating latent variables, $SD^2$ directly disentangles the observed features into three underlying factors by introducing causal mechanisms, which is more efficient and effective. In contrast to preceding causal disentanglement methods, our approach facilitates the generation of mutually independent representations of underlying factors without the need for intricate mutual information estimators. Instead, we fundamentally minimize the mutual information between the representations of underlying factors without explicit estimation, which will be elaborated in Section \ref{method}.
\section{Problem Setup}
\label{setup}
As shown in Figure \ref{causal graph}, we have observed pre-treatment features $X$, treatment $T$, and outcome $Y$ in the dataset $\mathcal{D}$. $X$ is composed of three types of underlying factors: 
\begin{itemize}
    \item \textbf{Instrumental variable} $Z$ that only directly influences $T$.
    \item \textbf{Confounders} $C$ that directly influences both $T$ and $Y$.
    \item \textbf{Adjustable variable} $A$ that only directly influences $Y$.
\end{itemize}   

Besides these, there are some \textbf{Unobserved confounders} $U$ that impede the counterfactual prediction. $Z, C, A, U$ in this paper are exogenous variables.

According to the d-separation theory \citep{pearl2000models}, $Z$, $C$ and $A$ are mutually independent due to the collider structures $Z$ $\rightarrow$ $T$ $\leftarrow$ $C$ and $C$ $\rightarrow$ $Y$ $\leftarrow$ $A$. Thus, encouraging independence among the representations of these underlying factors is essential for advancing their separation. In this paper, we aim to first disentangle mutually independent representations of $Z$, $C$, and $A$ from $X$, bypassing the explicit estimation of mutual information through theoretical analysis, then propose a unified framework to tackle confounding bias caused by $C$ and $U$ simultaneously. 

Given the above definitions, we present the formal definitions of counterfactual prediction problem, followed by an essential assumption for \textbf{identification} of causal effects adopted in this paper.

\begin{definition}
    \textbf{Counterfactual Prediction} refers to predicting what the outcome would have been for an individual under an intervention $t$ \citep{pearl2000models}, i.e.,
        \begin{equation}
        g(t, X) = E[Y | do(T=t), X].
        \label{counterfactual}
        \end{equation}
    \label{def:counter}
\end{definition}
where \textit{do} refers to do-operator proposed by \citet{pearl2000models}.

Assuming successful separation of $Z$, $C$, and $A$, the sufficient identification results for causal effects under the additive noise assumption in instrumental variable regression were developed by \citet{angrist1996identification}.

\begin{assumption} 
    \textbf{Additive Noise Assumption:} the noise from unmeasured confounders $U$ is added to the outcomes $Y$, i.e., 
    \begin{equation}
        Y = g(t,X) + U.
    \label{additive}
    \end{equation}

\label{assump:additive}
\end{assumption}
\section{Methodology}
\label{method}


\subsection{Theoretical Foundation of Independentizing Underlying Factors}
\label{theory}
\begin{figure}[t!]
\centering{\includegraphics[width=0.7\columnwidth]{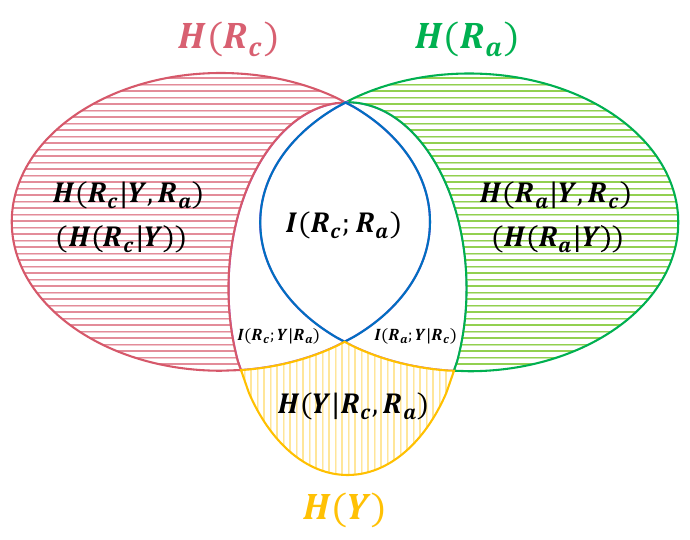}}
\caption{A motivating Venn diagram of mutual information between $R_c$, $R_a$ and $Y$ during training phase.   }
\label{venn}
\end{figure}

To get the mutually independent representations of underlying factors, an intuitive thought is to minimize the mutual information between the representation of $Z$, $C$ and $A$, i.e., $I(R_a; R_c)$  \footnote{$R_a$ and $R_c$ differ from $A$ and $C$ in Figure \ref{causal graph}. Our optimization aims to bring $R_a$ and $R_c$ to the states where $A$ and $C$ can be represented. Initially, $R_a$ and $R_c$ are entangled with non-zero mutual information and thus do not follow the causal relations in Figure \ref{causal graph}, as depicted in Figure \ref{venn}. We introduce constraints to reduce their MI and achieve independence.} and $I(R_z; R_c)$, where $I(\cdot)$ represents the function of mutual information, during training. However, it is hard due to the notorious difficulty in estimating mutual information in high dimensions, especially for latent embedding optimization \citep{belghazi2018mutual,poole2019variational}. To circumvent this challenge, we decompose the mutual information between $R_a$ and $R_c$ by leveraging the chain rule of mutual information:
\begin{equation}
I({R_a};{R_c}) = I(Y;{R_c}) + I({R_a};{R_c} \mid Y) - I({R_c};Y \mid {R_a}).
\label{chain}
\end{equation}
We further inspect the term $I({R_a};{R_c} \mid Y)$ in Eq \eqref{chain}. Based on the definition of mutual information, we have, 
\begin{equation}
I({R_a};{R_c} \mid Y) = H({R_a} \mid Y) - H({R_a} \mid Y, {R_c}),
\label{chain_1}
\end{equation}
where $H(\cdot)$ represents the function of entropy. Intuitively, this conditional mutual information measures the information contained in $R_a$ that is related to $R_c$ but unrelated to $Y$. As shown in Figure \ref{causal graph}, the only connection between $C$ and $A$ is that they determine $Y$ jointly with $T$. If we set up two prediction models from $R_a$ and $R_c$ to Y, respectively, then the mutual information between $R_a$ and $R_c$ is all related to $Y$ during the training phase. We draw the Venn diagram of mutual information between $R_a$, $R_c$ and $Y$ as shown in Figure \ref{venn}, according to which we have,
\begin{equation}
H({R_a} \mid Y) = H({R_a} \mid Y, {R_c}).
\label{chain_2}
\end{equation}
With Eq \eqref{chain}, \eqref{chain_1} and \eqref{chain_2}, we have,
\begin{equation}
I({R_a};{R_c}) = I(Y;{R_c}) - I({R_c};Y \mid {R_a}).
\label{chain_3}
\end{equation}
Eq \eqref{chain_3} transforms the mutual information between two training high-dimensional representations into the subtraction of two mutual information estimators with known labels. To further simplify it, we introduce the following theory:
\begin{theorem}
\label{theorem_1}
Minimizing the mutual information between $R_a$ and $R_c$ is equivalent to:
\begin{equation}
\min\, H(Y) - H(Y \mid R_c) - H(Y \mid R_a) + H(Y \mid R_c, R_a).
\label{chain_4}
\end{equation}
\end{theorem}
To find a tractable solution to Eq \eqref{chain_4}, we derive the following Corollary:
\begin{corollary}
\label{corollary_1}
One of sufficient conditions of minimizing $I({R_a};{R_c})$ is minimizing the following conditions together:
\begin{subnumcases}
    {\min\, \mathcal{L}_{c}^{a} = }
    {D}_{KL}[\mathcal{P}_{Y}\Vert \mathcal{P}^{R_a}_{Y} ]\label{7_1}\\
    {D}_{KL}[\mathcal{P}^{R_a}_{Y}\Vert \mathcal{P}^{R_c}_{Y} ]\label{7_2},
\end{subnumcases} 
where $\mathcal{P}^{R_a}_{Y}=p(Y \mid {R_a})$, $\mathcal{P}^{R_c}_{Y}=p(Y\mid{R_c})$ represent the predicted distributions of $Y$, $\mathcal{P}_{Y}=p(Y)$ represents the real distribution of $Y$. $D_{KL}$ denotes the KL-divergence.
\end{corollary}
Detailed proof and formal assertions of \textbf{Theo.\ref{theorem_1}} and \textbf{Corol.\ref{corollary_1}} can be found in the \textbf{Appendix \ref{4.2} and \ref{4.3}}. Similarly, we can transform minimizing $I(R_z; R_c)$ into following:
\begin{subnumcases}
    {\min\, \mathcal{L}_{c}^{z} = }
    {D}_{KL}[\mathcal{P}_{T}\Vert \mathcal{P}^{R_z}_{T} ]\label{8_1}\\
    {D}_{KL}[\mathcal{P}^{R_z}_{T}\Vert \mathcal{P}^{R_c}_{T} ]\label{8_2},
\end{subnumcases}
where $\mathcal{P}^{R_a}_{T}=p(T \mid {R_a})$, $\mathcal{P}^{R_z}_{T}=p(T\mid{R_z})$ represent the predicted distributions of $T$, $\mathcal{P}_{T}=p(T)$ represents the real distribution of $T$.

\textbf{Remark:} Technically, the optimization of Eq \eqref{7_1},\eqref{7_2} can be realized by training prediction networks about $Y$. Additionally, we can set up a deep prediction network from $T$, $A$, $C$ to $Y$ and use its supervision information to guide the training of the shallow prediction models. From this perspective, our approach is essentially a self-distillation model. Therefore, we name our method as \textbf{S}elf-\textbf{D}istilled \textbf{D}isentanglement, i.e., $SD^2$.

\subsection{Self-Distilled Disentanglement Framework}
\label{framework_section}
With the theory put forward in section \ref{theory}, we propose a self-distilled framework to disentangle different mutually independent underlying factors. To clarify further, we take the distillation unit for minimizing $\mathcal{L}_{c}^{z}$ to illustrate how we employ different sources of supervision information to directly minimize mutual information without explicitly estimating it.

\begin{figure}[t!]
\centering{\includegraphics[width=0.9\columnwidth]{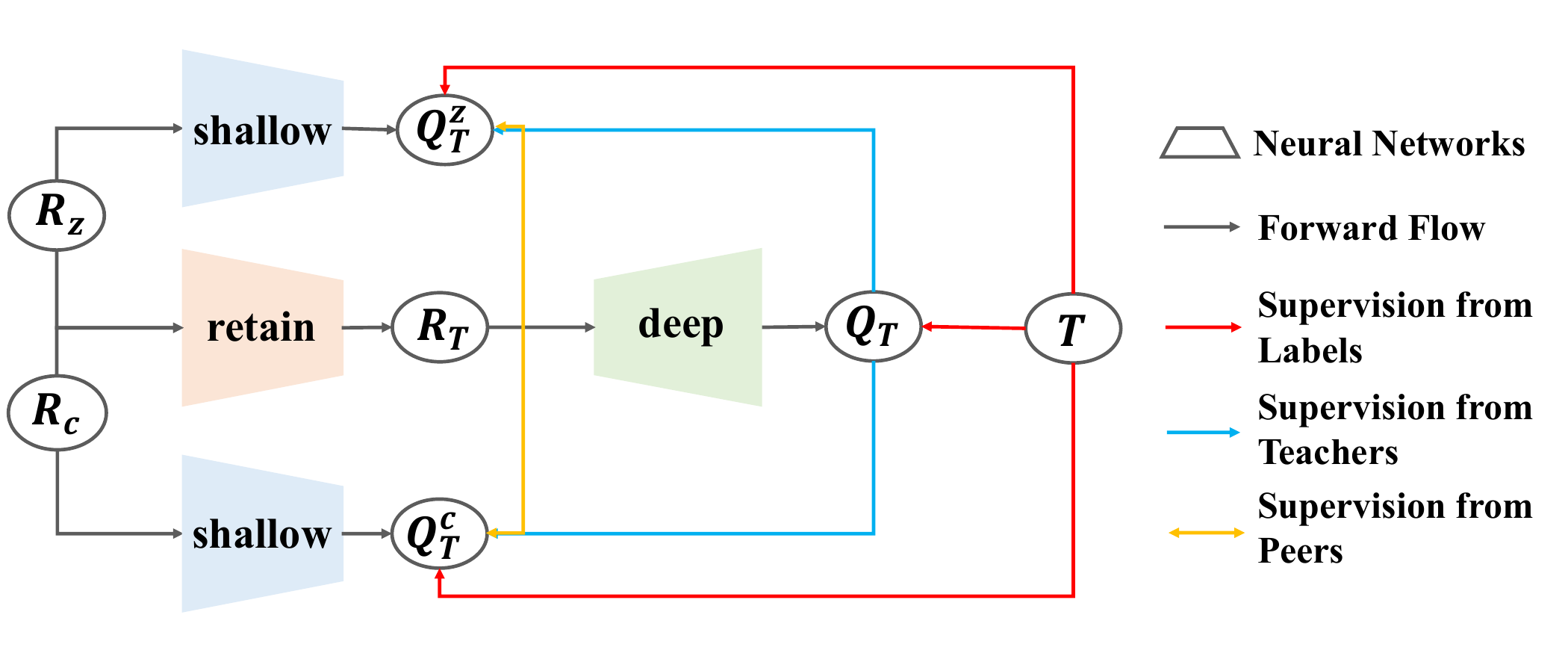}}
\caption{Self-distillation unit for minimizing $\mathcal{L}_{c}^{z}$.   }
\label{framework}
\end{figure}

As shown in Figure \ref{framework}, \textbf{Retain} network represents the neural network for retaining the information from both $Z$ and $C$. \textbf{Deep} networks and \textbf{Shallow} networks are named based on their relative proximity to $R_z$ and $R_c$. We set up two Shallow prediction networks from $R_z$ and $R_c$ to $T$, respectively. In addition, to ensure $R_z$ and $R_c$ grasp sufficient information for predicting $T$, we set up a Retain network to concatenate $R_z$ and $R_c$ and store joint information of them for input into a Deep prediction network. To minimize $\mathcal{L}_{c}^{z}$, we deploy distinct sources of supervision information from: 

(1) \textbf{Labels;} We use $T$ directly to guide the training of deep prediction networks by minimizing the prediction loss $ L(Q_T, T)$. For shallow prediction networks, we reduce ${D}_{KL}[\mathcal{P}_{T}\Vert \mathcal{P}^{R_z}_{T} ]$ and ${D}_{KL}[\mathcal{P}_{T}\Vert \mathcal{P}^{R_c}_{T} ]$ by minimizing prediction loss $ L(Q_T^z, T)$ and $ L(Q_T^c, T)$.\footnote{In practice, only minimizing Eq \eqref{8_1} and \eqref{8_2} makes convergence challenging, resulting in unsatisfactory performance. We speculate that this is due to a lack of sufficient supervised information guiding the updating direction of the prediction network for $R_c$. Therefore, we introduce the minimization of ${D}_{KL}[\mathcal{P}_{T}\Vert \mathcal{P}^{R_c}_{T} ]$ (corresponding loss function $ L(Q_T^c, T)$ to expedite loss convergence. }

(2) \textbf{Teachers;} We regard retain and deep prediction networks as a teacher model, which can convey the learned knowledge to help the training of shallow prediction networks. That is, minimizing KL-divergence loss $L(Q_T^z, Q_T)$ and $ L(Q_T^c, Q_T)$.

(3) \textbf{Peers;} We diminish the KL-divergence between the distributions of the outputs from the two shallow prediction networks, i.e., $ L(Q_T^c, Q_T^z)$. ${D}_{KL}[\mathcal{P}^{R_z}_{T}\Vert \mathcal{P}^{R_c}_{T} ]$ are consequently minimized.
Similarly, we can minimize $\mathcal{L}_{c}^{a}$ by establishing corresponding distillation units according to \textbf{Corol.\ref{corollary_1}}. In addition, \citet{hassanpour2019learning} propose to separate $A$ from $X$ based on $A\perp T$. Specifically, for binary treatment setting, they minimize the distribution discrepancy of representations of $A$ between the treatment group ($T$ = 1) and the control group ($T$ = 0). The related loss function can be defined as follows:
\begin{equation}
\min\, \mathcal{L}_{a}= disc({\{R_a^i\}_{i:t_i=0}},{\{R_a^i\}_{i:t_i=1}}),
\label{R_a}
\end{equation}
where function $disc(\cdot)$ represents the distribution discrepancy of $R_a$ between the treatment and control groups while \textit{i} refers to the \textit{i-th} individual. We also provide continuous version of $\mathcal{L}_{a}$ in the 
\textbf{Appendix \ref{conti}.}

\textbf{Mitigating Confounding Bias:} With disentangled confounders $C$, \textbf{the confounding bias induced by $C$} can be alleviated by re-weighting the factual loss $ L(Q_Y, Y)$ with the context-aware importance sampling weights ${\omega}_{i}$ \citep{hassanpour2019learning} for each individual $i$. Besides, $Q_T$, the output of the deep prediction network for $T$, can be employed to regress $Y$, which helps to mitigate \textbf{the confounding bias caused by unobserved confounders $U$} \citep{hartford2017deep}.

The loss function of $SD^2$ is thus devised as follows, where $L$ can be any functions measuring the differences between two items. $\alpha$, $\beta$, $\gamma$, $\delta$ are adjustable hyper-parameters. The $l_{2}$ regularization loss ${{\parallel}\cdot {\parallel}}_{2}$ on model weights $W$ is to avoid over-fitting.
\begin{equation}
\begin{split}
    \mathcal{L}_{SD^2}= &\underbrace{\sum_{i} {\omega}_{i} L({Q_Y}_{i},{Y}_{i})+\alpha L(Q_T,T)}_{factual\; loss}+ \\
&\underbrace{\beta \mathcal{L}_{a} + \gamma ( \mathcal{L}_{c}^{a}+ \mathcal{L}_{c}^{z})}_{disentanglement\; loss}+
\underbrace{\delta {{\parallel}W{\parallel}}_{2}}_{regularization\; loss}.
\label{loss}
\end{split}
\end{equation}

\begin{table*}[t]
\caption{Performance comparison of bias of ATE between $VSD^2$ and the SOTA baselines on the synthetic datasets ($mv$-$mz$-$mc$-$ma$-$mu$) and Twins datasets (Twins-$mv$-$mx$-$mu$). Bold/Underline indicates the method with the best/second best performance.  }
\vskip 0.15in
\begin{scriptsize}
\begin{center}
    \centering
    \renewcommand\arraystretch{1.6}
    \setlength{\tabcolsep}{4pt}
    \begin{tabular}{c c c c c c c c c}
    \hline
    &\multicolumn{4}{c}{\textbf{Within-Sample}}&\multicolumn{4}{c}{\textbf{Out-of-Sample}} \\ \hline
    \textbf{Method}&\textbf{0-4-4-2-2}&\textbf{2-4-4-2-2}&\textbf{0-4-4-2-10}&\textbf{0-6-2-2-2}& \textbf{0-4-4-2-2} & \textbf{2-4-4-2-2} & \textbf{0-4-4-2-10} & \textbf{0-6-2-2-2} \\ \hline    
    \textbf{DirectRep}&0.037(0.023)&0.034(0.009)&0.060(0.014)&0.043(0.025)& 0.034(0.024) & 0.034(0.013) & 0.060(0.015) &0.037(0.024)\\
    \textbf{CFR}&\underline{0.031(0.014)}&\underline{0.032(0.017)}&0.056(0.017)&0.060(0.019)&\underline{0.027(0.016)} & \underline{0.032(0.021)} & 0.031(0.024) & 0.054(0.018) \\ 
    \textbf{DRCFR}&0.058(0.018)&0.050(0.025)&0.071(0.020)&0.066(0.015)& 0.053(0.016) & 0.050(0.023) & 0.071(0.020) & 0.059(0.018)\\ 
    \textbf{DFL}&3.696(0.034)&3.702(0.045)& 4.055(0.041) & 3.450(0.030)& 6.796(0.056) & 6.750(0.086) & 7.398(0.051) & 6.393(0.063) \\ \hline
    \textbf{DeepIV-Log}&0.569(0.024)&0.567(0.011)&0.601(0.011)&0.531(0.019)& 0.572(0.028) & 0.567(0.016) & 0.601(0.011) &0.537(0.021) \\ 
    \textbf{DeepIV-Gmm}&0.466(0.012)&0.387(0.021)&0.518(0.012)&0.437(0.007)& 0.469(0.008) & 0.387(0.021) & 0.517(0.009) &0.442(0.004) \\ 
    \textbf{DFIV}&3.947(0.108)&3.810(0.041)&4.400(0.134)&3.644(0.103)& 7.047(0.131) & 6.856(0.093) & 7.749(0.125) & 6.585(0.130) \\ 
    \textbf{OneSIV}&0.504(0.008)&0.441(0.063)&0.569(0.013)& 0.478(0.012)& 0.507(0.009) & 0.441(0.064) & 0.569(0.015) & 0.484(0.014)\\ 
    \textbf{CBIV}&0.063(0.025)&0.065(0.032)&\underline{0.049(0.022)}&\underline{0.033(0.023)}& 0.059(0.024) & 0.067(0.030) & \underline{0.049(0.023)} &\underline{0.030(0.017)}\\ \hline
    \textbf{Ours}&\textbf{0.010(0.008)}&\textbf{0.017(0.013)}&\textbf{0.029(0.019)}&\textbf{0.014(0.013)}& \textbf{0.012(0.008)} & \textbf{0.022(0.017)} & \textbf{0.029(0.018)} & \textbf{0.013(0.010)} \\ \hline 
    \hline
    &\multicolumn{4}{c}{\textbf{Within-Sample}}&\multicolumn{4}{c}{\textbf{Out-of-Sample}} \\ \hline
    \textbf{Method}&\textbf{Twins-0-16-8}&\textbf{Twins-4-16-8}&\textbf{Twins-0-16-12}&\textbf{Twins-0-20-8}&\textbf{Twins-0-16-8}&\textbf{Twins-4-16-8}&\textbf{Twins-0-16-12}&\textbf{Twins-0-20-8}\\ \hline
    \textbf{DirectRep}&0.015(0.007)&0.009(0.008)&0.018(0.012)&0.012(0.007)&0.023(0.011)&0.014(0.013)&0.023(0.016)&0.014(0.010)\\ 
    \textbf{CFR}&\underline{0.011(0.010)}&0.012(0.008)&0.017(0.014)&\underline{0.010(0.011)}&\underline{0.016(0.008)}&0.019(0.010)&0.025(0.019)&0.019(0.014)\\ 
    \textbf{DRCFR}&0.015(0.016)&\underline{0.008(0.005)}&\underline{0.016(0.017)}&0.024(0.017)&0.020(0.019)&0.018(0.007)&0.020(0.009)&0.028(0.018)\\ 
    \textbf{DFL}&0.173(0.018)&0.181(0.014)& 0.169(0.020) & 0.175(0.019)&0.243(0.136)&0.252(0.140)& 0.240(0.141) & 0.245(0.145) \\ \hline
    \textbf{DeepIV-Log}&0.019(0.010)&0.017(0.016)&0.022(0.011)&0.012(0.007)&0.028(0.016)&0.026(0.019)&0.028(0.018)&0.018(0.011) \\ 
    \textbf{DeepIV-Gmm}&0.022(0.004)&0.016(0.004)&0.022(0.003)&0.018(0.004)&0.017(0.011)&0.013(0.009)&\underline{0.017(0.011)}&\underline{0.013(0.011)} \\ 
    \textbf{DFIV}&0.186(0.014)&0.185(0.015)&0.186(0.014)&0.163(0.023)&0.257(0.145)&0.255(0.148)&0.256(0.144)&0.227(0.151) \\ 
    \textbf{OneSIV}&0.015(0.011)&\underline{0.008(0.005)}&0.030(0.017)&0.017(0.006)&0.016(0.014)&\underline{0.011(0.007)}&0.025(0.020)&0.013(0.010) \\ 
    \textbf{CBIV}&0.024(0.016)&0.057(0.012)&0.051(0.038)&0.037(0.019)&0.033(0.016)&0.063(0.017)&0.057(0.039)&0.043(0.022)\\ \hline    \textbf{Ours}&\textbf{0.008(0.006)}&\textbf{0.007(0.004)}&\textbf{0.007(0.006)}&\textbf{0.008(0.006)}&\textbf{0.012(0.007)}&\textbf{0.011(0.006)}&\textbf{0.012(0.006)}&\textbf{0.011(0.009)} \\ \hline
    
    \end{tabular}
\end{center}
\end{scriptsize}
\label{result1}
\end{table*}
\section{Experiments}
\label{exp}

\subsection{Benchmarks }
Due to the absence of counterfactual outcomes in reality, it is challenging to conduct counterfactual prediction on real-world datasets. Previous works \citep{shalit2017estimating,10027640,wu2020learning} synthesize datasets or transform real-world datasets. In this paper, we have conducted multiple experiments on a series of synthetic datasets and a real-world dataset, Twins. The source code of our algorithm is available on \textbf{GitHub}\footnote{\url{https://github.com/XinshuLI2022/SDD}}.

\subsubsection{Simulated Datasets}
\textbf{\textit{Binary Scenario}}: 
Similar to \citet{hassanpour2019learning}\footnote{The difference between our data generation process and that of \citet{hassanpour2019learning} lies in the introduction of unobserved confounders $U$, in the generation processes of $T$ and $Y$. Additionally, we categorize IVs into observed and unobserved ones, facilitating comparisons with various baseline algorithms.}, we generate the \textit{synthetic} datasets according to the following steps:
\begin{itemize}
    \item For $K$ in {$Z$, $C$, $A$, $V$, $U$}, sample $K$ from $\mathcal{N}(0, {I}_{mk})$, where ${I}_{mk}$ denotes $mk$ degree identity matrix. Concatenate $Z$, $C$, and $A$ to constitute the observed covariates matrix $X$. $V$ represents the observed IVs, the dimension of which could be set to 0. The setting of $V$ is mainly for the comparison of IV-based methods. $U$ represents the unobserved confounders.
    \item Sample treatment variables $T$ as following: 
    \begin{equation}
    \begin{split}
        {T_i}\sim&Bern(Sigmoid(\\
        &\sum\limits^{mz}\limits_{i=1}{Z}_{i}+\sum\limits^{mc}\limits_{i=1}{C}_{i}+\sum\limits^{mv}\limits_{i=1}{V}_{i}+\sum\limits^{mu}\limits_{i=1}{U}_{i})),
    \end{split}
    \label{sample_t}
    \end{equation}
    where $mz$, $mc$, $mv$, $mu$ represent the dimension of $Z$, $C$, $V$, $U$ respectively.
    \item Sample outcome variables $Y$ as following: 
    \begin{equation}
    \begin{split}
        {Y_i}\sim&Bern(Sigmoid((\\
        &\frac{Ti}{ma+mc+mu}{\sum\limits^{ma}\limits_{i=1}}{A_i}^{2}+{\sum\limits^{mc}\limits_{i=1}}{C_i}^{2}+{\sum\limits^{mu}\limits_{i=1}}{U_i}^{2})+\\
        &(\frac{1-Ti}{ma+mc+mu}{\sum\limits^{ma}\limits_{i=1}}{A}_{i}+{\sum\limits^{mc}\limits_{i=1}}{C}_{i}+{\sum\limits^{mu}\limits_{i=1}}{U}_{i}))),
    \end{split}
    \label{sample_y}
    \end{equation}
    where $ma$ represents the dimension of $A$.    
\end{itemize}

\textbf{\textit{Continuous Scenario}}: Following the work of \citet{hartford2017deep,wu2022instrumental}, we use the \textit{Demand} datasets to evaluate the performance of $SD^2$ on the continuous scenario with the same data generation process described in detail in \citet{wu2022instrumental}. 

\subsubsection{Real-World Datasets}
\textit{Twins}: 
The Twins dataset comprises the records of the twins who were born between 1989-1991 in the USA \citep{almond2005costs}. We select 5271 records of same-sex twins who weighed less than 2kg and had no missing features in the records. The heavier twin in twins is assigned with $T$ = 1 and the lighter one with $T$ = 0. The mortality after one year is the observed outcome. It is reasonable that only part of the features determine the outcome. Therefore, we randomly generate $mv$-dimension $V$ and choose some features $M$ as a combination of $Z$, $C$ and $U$ to create $T$ according to the policy in Eq.\eqref{sample_t}. The rest features $R$ will include $A$ and some noise naturally. We hide some features in $M$ during training to create $U$ and treat the rest features in $M$ with $R$ as $X$. 

During training, only $X$, $V$, $T$ and $Y$ will be accessible under any scenario.

\begin{figure*}[htbp]
\centerline{\includegraphics[width=\textwidth]{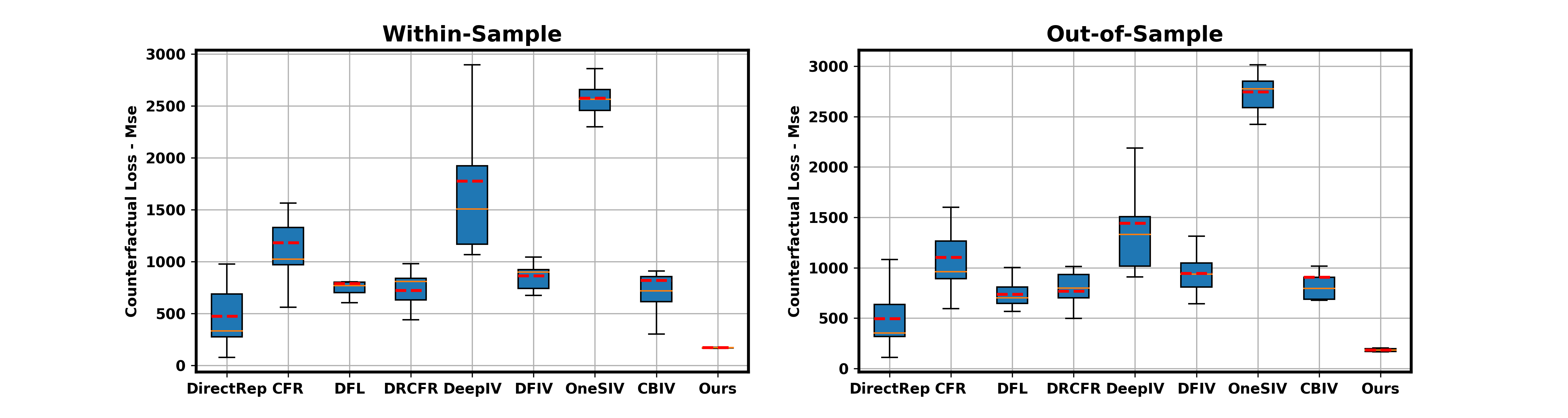}}
\caption{Experimental results under continuous scenario on Demand-0-1. Among all methods, $VSD^2$ achieves the best and most stable results.    }
\label{demand}
\end{figure*}
 \begin{table}
\caption{Performance comparison of MSE between $SD^2$ and the SOTA baselines on the Demands datasets (Demands-$\alpha$-$\beta$). Bold/Underline indicates the method with the best/second-best performance.  }
\vskip 0.15in
\begin{scriptsize}
\begin{center}
    
    \renewcommand\arraystretch{1.6}
    \setlength{\tabcolsep}{7pt}
    \centering
    \begin{tabular}{c c c c}
    \hline
    &\multicolumn{3}{c}{\textbf{Within-Sample}}\\ \hline
    \textbf{Method}&\textbf{Demands-0-1}&\textbf{Demands-0-5}&\textbf{Demands-5-1}\\ \hline    
    \textbf{DirectRep}&\underline{472.14(292.68)}&\underline{319.27(156.58)}&\underline{795.38(304.10)}\\
    \textbf{CFR}&1180.26(439.56)&340.12(175.82)&1891.91(79.78) \\ 
    \textbf{DFL}&786.92(122.35)&1074.71(142.78)& 874.78(103.71) \\ 
    \textbf{DRCFR}&720.51(195.09)&451.31(407.55)&809.04(94.98) \\ 
    \textbf{DEVAE}&\textgreater 10000&\textgreater 10000&\textgreater 10000\\ \hline
    \textbf{DeepIV-Gmm}&1774.71(757.98)&3429.79(438.65)&2618.72(775.20)\\ 
    \textbf{DFIV}&862.00(123.71)&1420.24(257.10)&909.25(137.05)\\ 
    \textbf{OneSIV}&2573.15(157.31)&\textgreater 10000&2352.69(125.16) \\ 
    \textbf{CBIV}&819.52(535.05)&410.42(241.71)&2542.58(442.26)\\ \hline\hline
    \textbf{Ours}&\textbf{170.55(6.51)}&\textbf{216.55(47.11)}&\textbf{171.07(6.41)}\\ \hline  
        &\multicolumn{3}{c}{\textbf{Out-of-Sample}} \\ \hline
    \textbf{Method}&\textbf{Demands-0-1}&\textbf{Demands-0-5}&\textbf{Demands-5-1}\\ \hline    
    \textbf{DirectRep}&\underline{494.47(287.92)} & \underline{303.22(144.85)} & \underline{768.15(278.24)} \\
    \textbf{CFR}&1103.00(362.45)&326.54(110.31) & 1877.60(119.71) \\ 
    \textbf{DFL} & 736.35(123.44)& 914.08(90.11) & 822.71(65.88) \\ 
    \textbf{DRCFR}&766.22(188.86)& 498.22(363.25) & 864.21(83.39) \\ 
    \textbf{DEVAE}&\textgreater 10000& \textgreater 10000 & \textgreater 10000 \\ \hline
    \textbf{DeepIV-Gmm}&904.60(618.41)& 2423.33(326.23)& 2405.33(675.40) \\ 
    \textbf{DFIV}&943.57(188.70)& 1213.27(388.08)& 1050.65(253.38) \\ 
    \textbf{OneSIV}& 2744.87(182.35)& 4243.51(596.10)& 2538.77(147.29) \\ 
    \textbf{CBIV}&2990.19(735.92)& 316.66(114.89) & 2958.70(614.26)\\ \hline
    \textbf{Ours}&\textbf{183.21(14.18)}& \textbf{199.44(14.60)} & \textbf{187.20(16.05)} \\ \hline  
    \end{tabular}
\end{center}
\end{scriptsize}
\label{demands}
\end{table}

\subsection{Comparison Methods and Metrics}
The \textbf{baselines} can be classified into two categories: 

\textbf{IV-based Methods}: DeepIV-Log and DeepIV-Gmm \citep{hartford2017deep}, DFIV \citep{xu2020learning}, OneSIV \citep{lin2019one}, CBIV \citep{wu2022instrumental}. These methods need well-predefined IVs $V$. When there is no $V$, i.e., $m_v$ equals 0, we use $X$ as $V$ for comparison. 

\textbf{Non-IV-based Methods}: DirectRep, CFR-Wass \citep{shalit2017estimating}, DFL \citep{xu2020learning}, DRCFR \citep{hassanpour2019learning}, CEVAE \citep{Louizos2017CausalEI}. The latter two also disentangle the hidden/latent factors from observed features.

\textbf{Metrics}: We use the absolute bias of Average Treatment Effect, i.e., $\epsilon ATE$, to evaluate the performance of different algorithms in the binary scenario. Formally, 
\begin{equation}
    \epsilon{ATE}=\mid{\frac{1}{N}[\sum\limits^N\limits_{i=1}(Y_i^1-Y_i^0)-\sum\limits^N\limits_{i=1}(\hat{Y_i^1}-\hat{Y_i^0})]}\mid,
    \label{ate}
\end{equation}

where ${Y_i}/\hat{Y_i}$ represents the factual/predicted potential outcome. For continuous scenarios, we take Mean Squared Error ($MSE$) as the evaluation metric. \textbf{The smaller $\epsilon ATE$ and $MSE$ are, the better the performance.}

\begin{figure}[t]
\centering
\subfigure[\scriptsize{Identification of Z with $VSD^2$}]{
\includegraphics[width=0.4\linewidth]{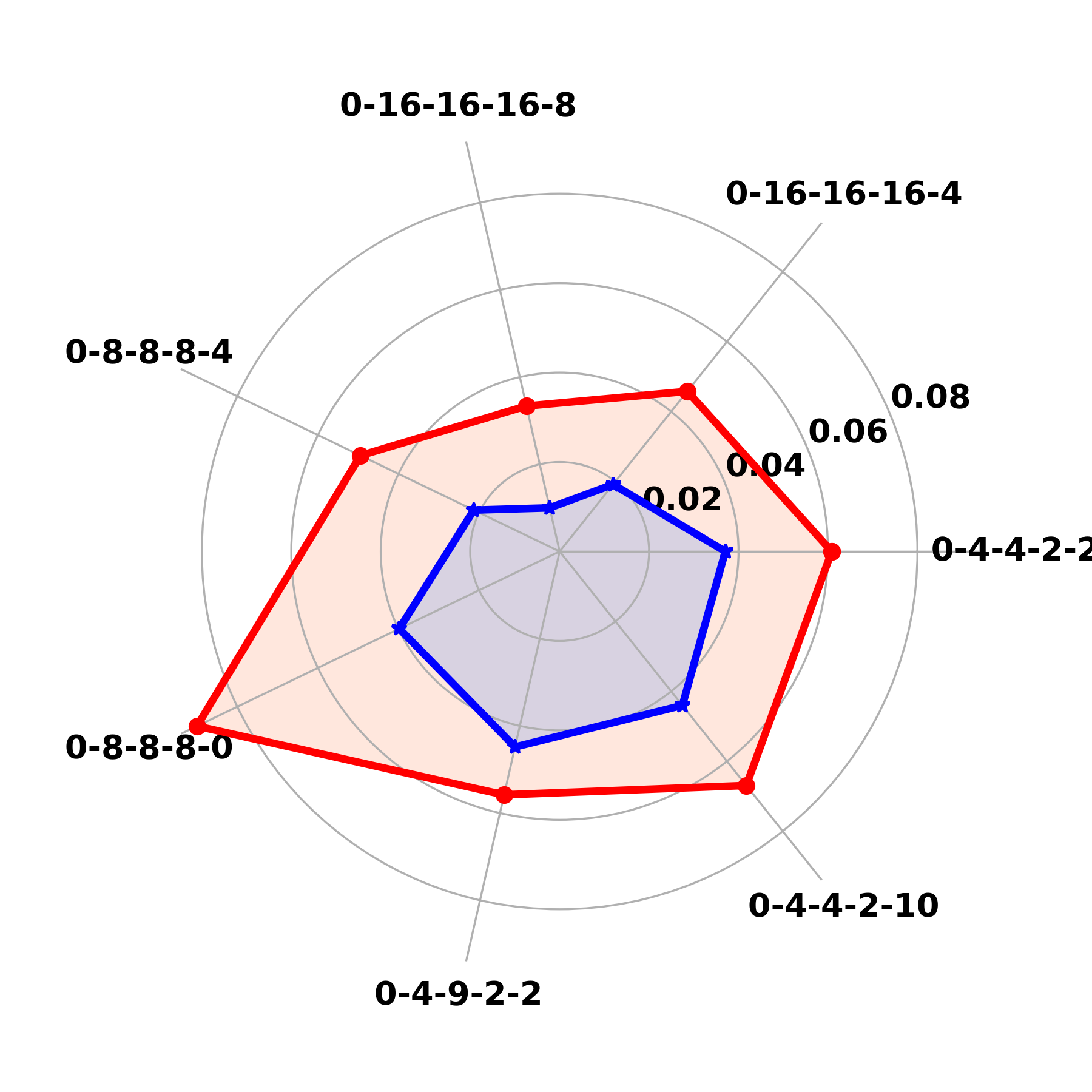}
\label{z_ours}
}
\subfigure[\scriptsize{Identification of Z with DRCFR}]{
\includegraphics[width=0.4\linewidth]{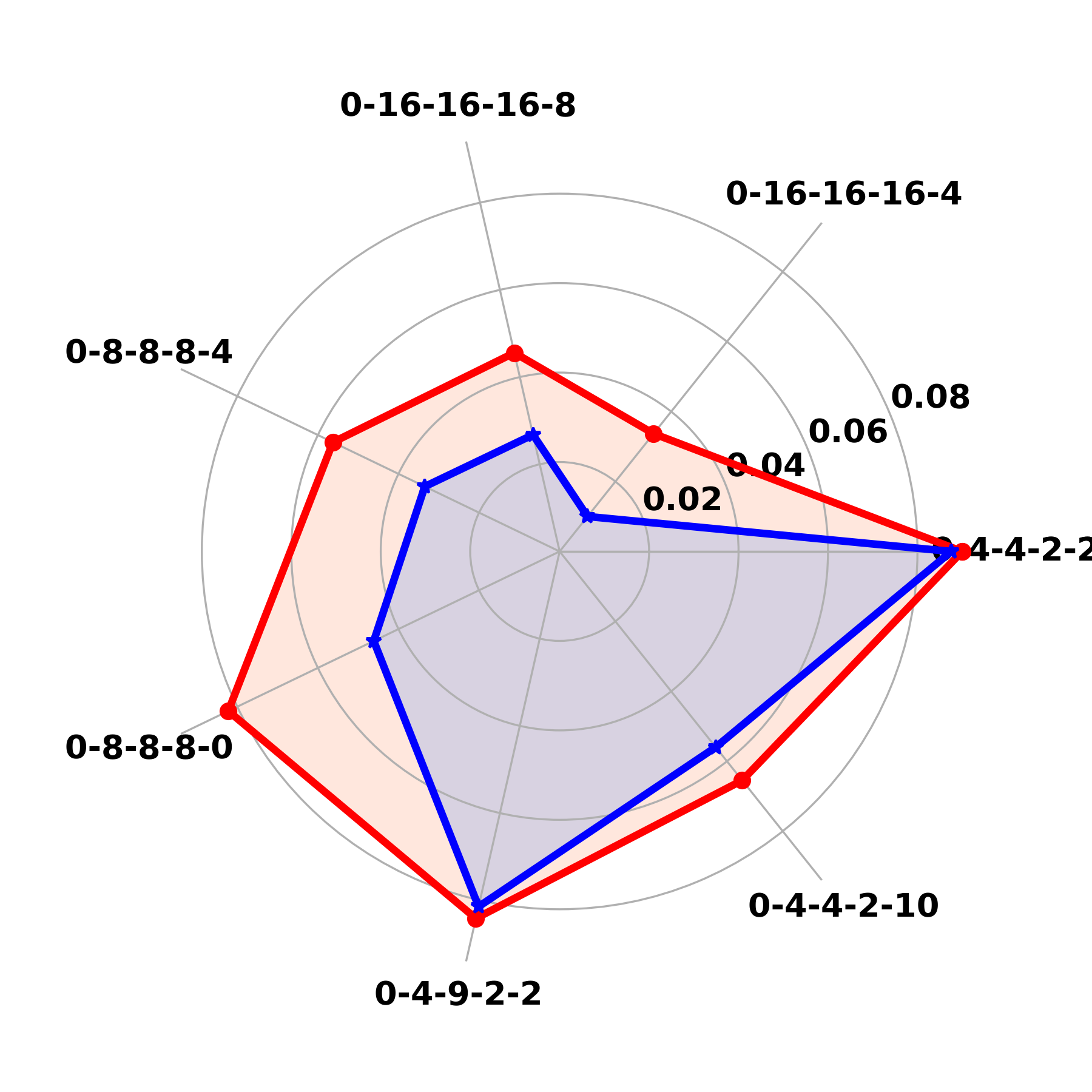}
\label{z_drcfr}
}
\quad
\subfigure[\scriptsize{Identification of C with $VSD^2$}]{
\includegraphics[width=0.4\linewidth]{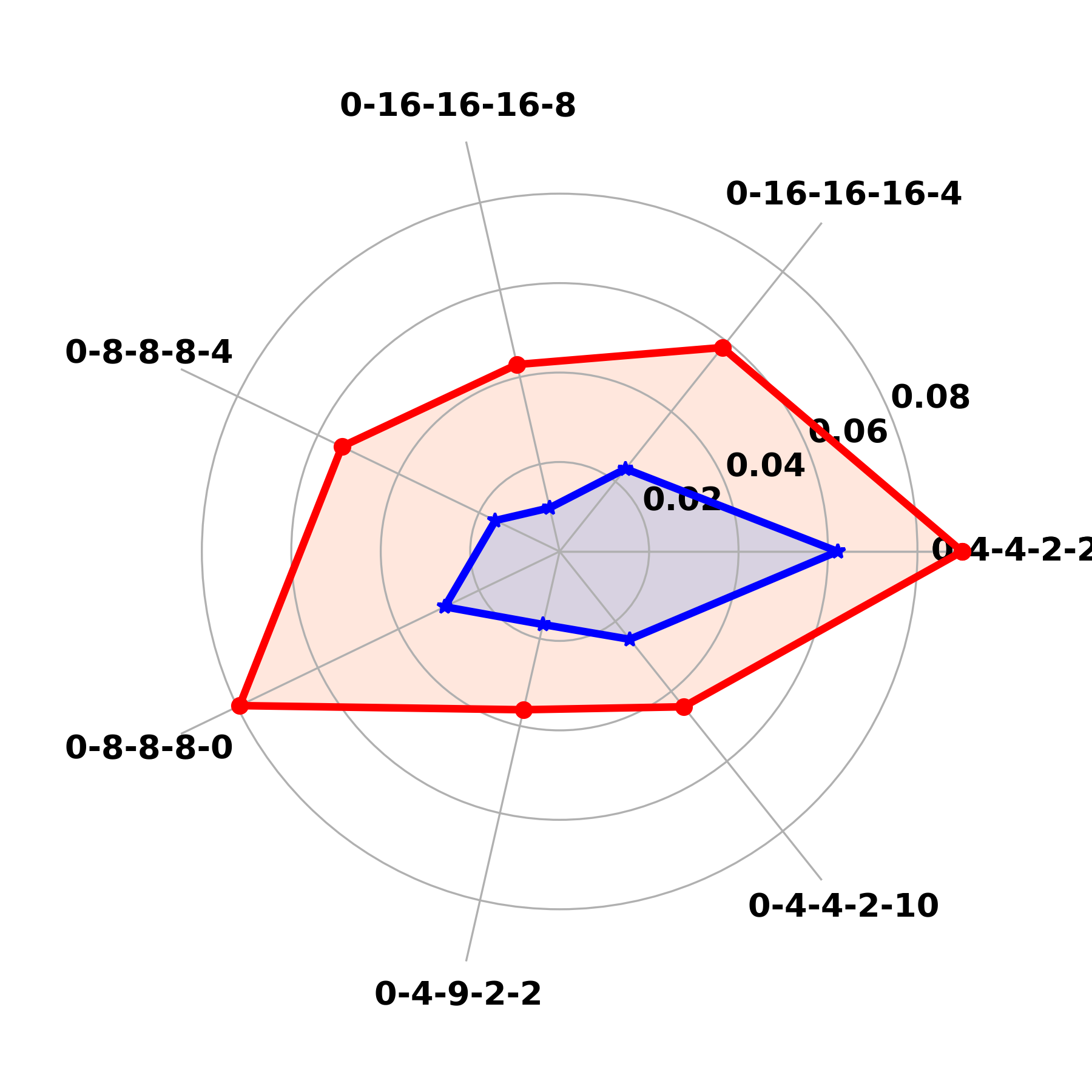}
\label{c_ours}
}
\subfigure[\scriptsize{Identification of C with DRCFR}]{
\includegraphics[width=0.4\linewidth]{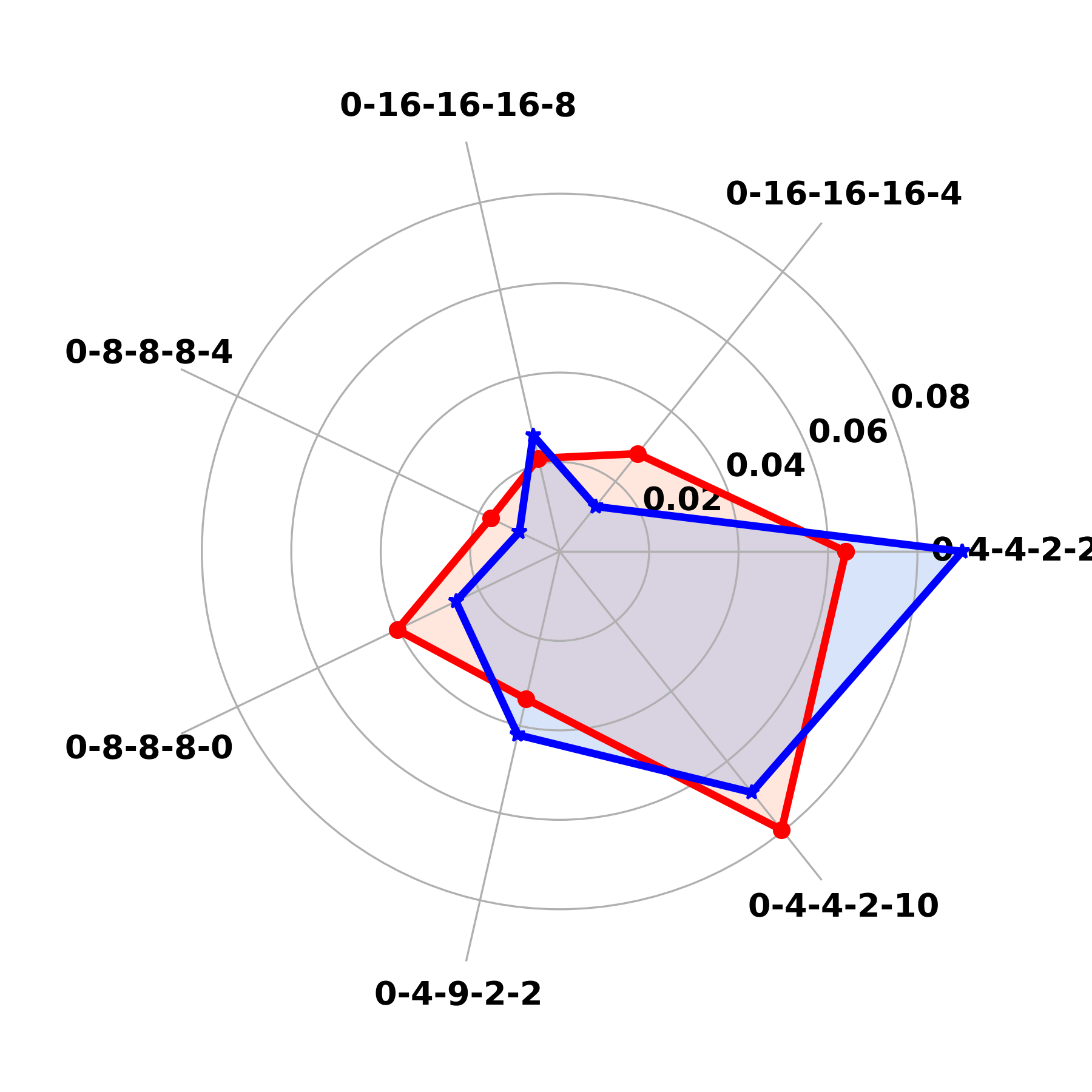}
\label{c_drcfr}
}
\quad
\subfigure[\scriptsize{Identification of A with $VSD^2$}]{
\includegraphics[width=0.4\linewidth]{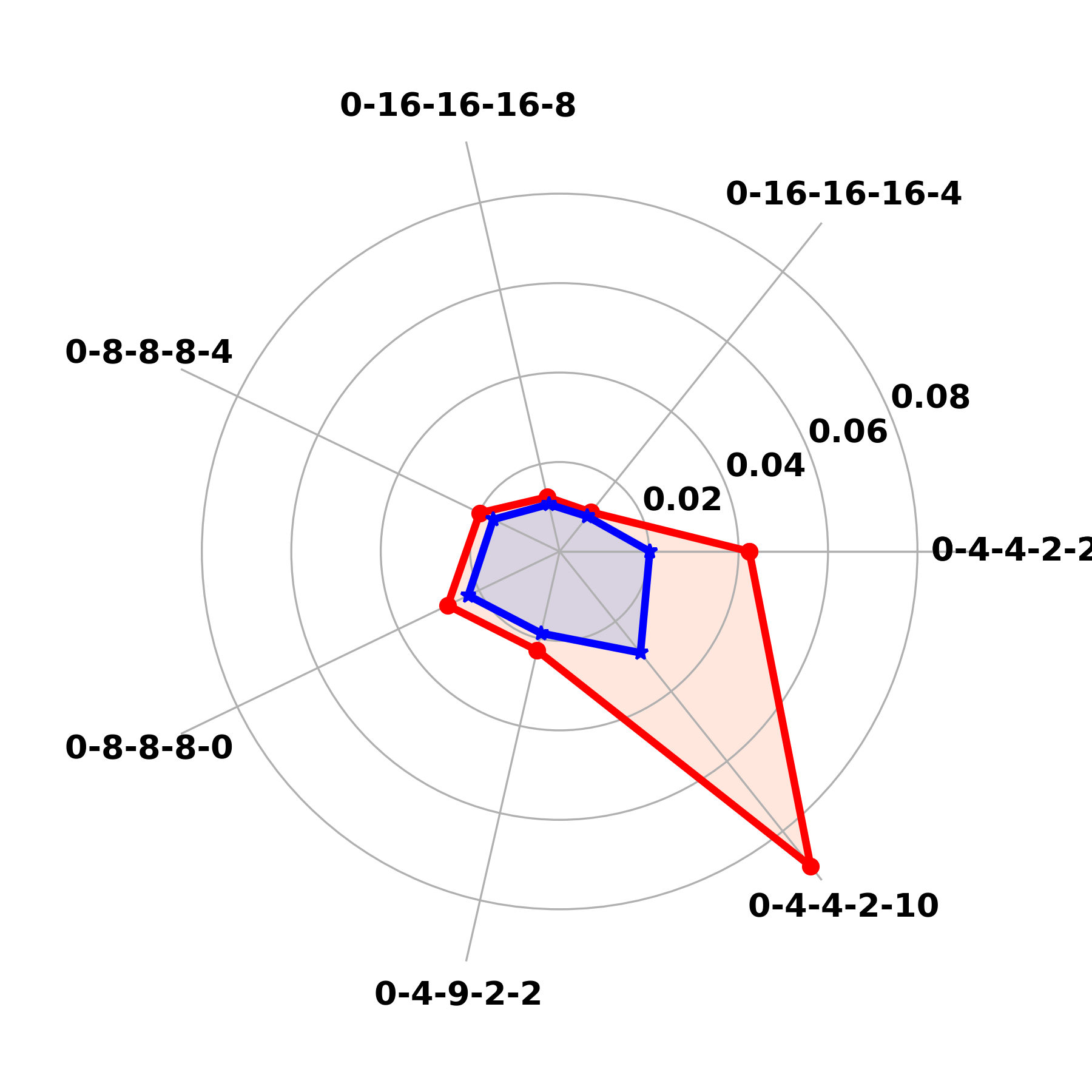}
\label{a_ours}
}
\subfigure[\scriptsize{Identification of A with DRCFR}]{
\includegraphics[width=0.4\linewidth]{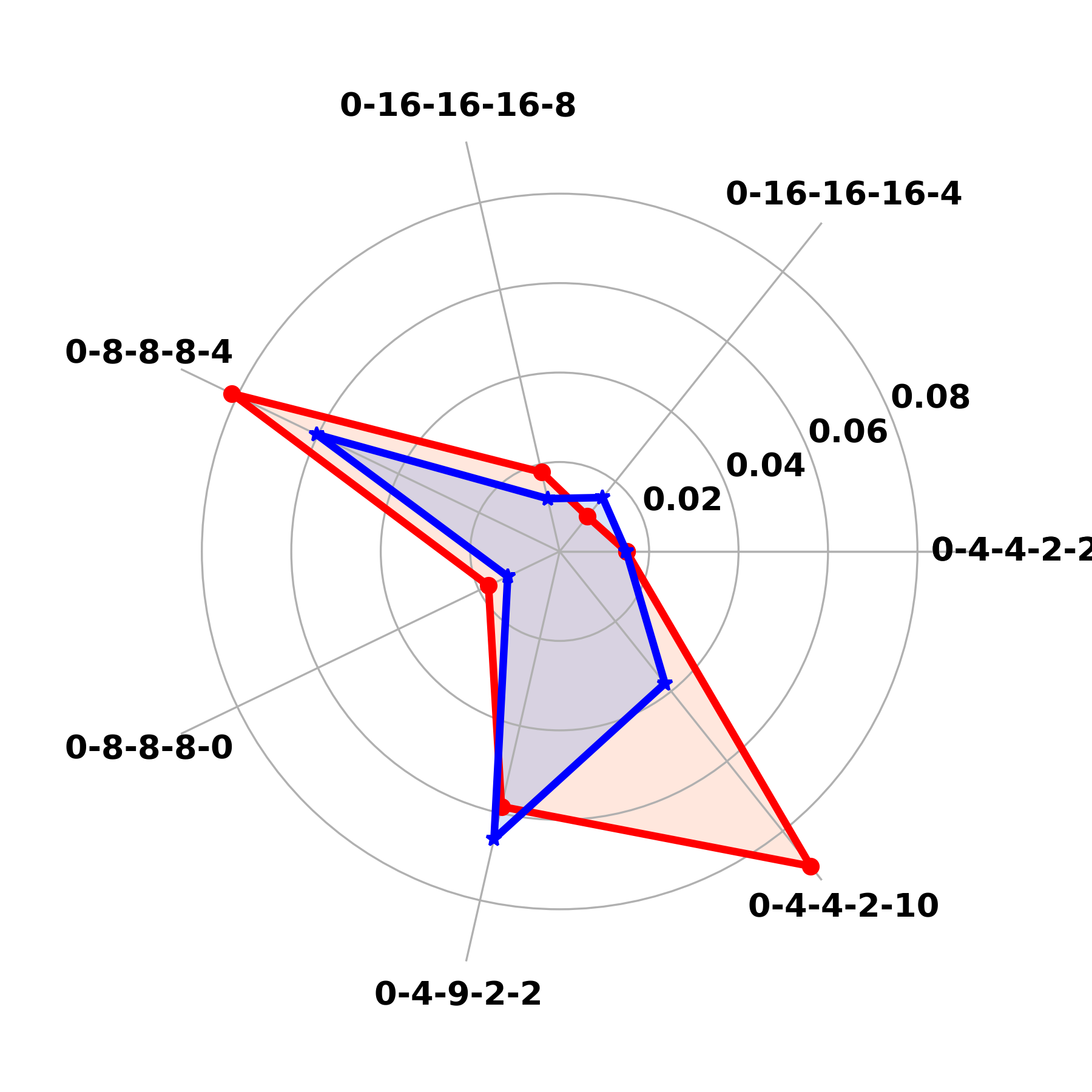}
\label{a_drcfr}
}
\caption{Radar charts that visualize the capability of $VSD^2$ and a classical causal disentangled learning baseline DRCFR. Every vertex on the polygons represents a synthetic dataset with setting $mv$-$mz$-$mc$-$ma$-$mu$. The red and blue denote the contribution of true variables and other variables to the decomposed representations. The results demonstrate our method achieves much better identification performance of all three underlying factors in all synthetic datasets compared with DRCFR.  }\label{visual}
\end{figure}

\subsection{Results}
\subsubsection{Comparison with the SOTA methods}
\label{syn_res}

\textbf{Binary Scenario}: Table \ref{result1} shows the performance of $SD^2$ on datasets with binary values. $m_v$-$m_z$-$m_c$-$m_a$-$m_u$ represents the data generated with $m_v$ predefined IVs $V$, $m_z$ underlying IVs $Z$, $m_c$ underlying confounders $C$, $m_a$ underlying adjustable variables $A$ and $m_u$ unobserved confounders $U$. During training, we can only observe $Z$, $C$ and $A$ as a whole $X$. Twins-$m_v$-$m_x$-$m_u$ denotes the Twins dataset with $m_v$ predefined IVs, $m_x$ observed pre-treatment covariates and $m_u$ unobserved confounders and IVs. For data setting with $m_{x}=16$ (Twins-$0$-$16$-$8$, Twins-$4$-$16$-$8$, Twins-$0$-$16$-$12$), there are 12 confounders and IVs, 4 adjustable variables and noises, while for Twins-$0$-$20$-$8$, we add 4 confounders and IVs in $X$. 

We generate 10000 samples for synthetic datasets for training, validation and testing sets, respectively, For the Twins datasets, we randomly choose 5271 samples and split the datasets with the ratio 63/27/10. We perform 10 replications and report the mean and standard deviation of the bias of ATE estimation. Taking the results in 0-4-4-2-2 as a base point for observation, we have the following findings: 
\begin{itemize}
    \item The results of almost all IV-based methods in 0-4-4-2-2 are worse than those in 2-4-4-2-2, demonstrating the performance of IV-based methods is highly dependent on the predefined instrumental variables.
    \item By comparing the results between 0-4-4-2-2 and 0-4-4-2-10, we find that when there are more unobserved confounders in the dataset, the performance of confounder balancing methods (such as CFR, DRCFR, and DFL) will be poorer, which indicates the necessity of controlling the unobserved confounders.
    \item If there are fewer confounders in observed features, as we compare 0-4-4-2-2 with 0-2-6-2-2, the performance of confounder balancing methods without disentanglement (such as CFR, DFL) gets impeded, implying the significance of precise confounder control by decomposing underlying factors.
    \item $SD^2$ achieves the best performance among all data settings and is far better than the second-best methods. It proves counterfactual prediction benefits from simultaneously controlling the underlying confounders in observational features and unobserved confounders. The \textit{\textbf{effectiveness}} of our disentanglement theory and hierarchical self-distillation framework is thus validated. The results in all Twins datasets are consistent with these findings.
\end{itemize}

\textbf{Continuous Scenario}: Following \citet{wu2022instrumental}, we use Demand-$\alpha$-$\beta$ to denote different data setting in Demands datasets. Demand-0-1 represents the original Demand dataset defined in \citet{hartford2017deep}. The $\alpha$ in Demand-$\alpha$-$\beta$ denotes the extra information from instrumental variables while the $\beta$ indicates the information increasing from instrumental variables and underlying confounders together on the basis of Demand-0-1. We only present the results on Demand-0-1 with box plots as shown in Figure \ref{demand}, where we omit the results of CEVAE as its $MSE$ is beyond 10000. 

The IV-based methods perform worse than the non-IV-based ones under the continuous scenario. This result indicates that confounding bias from the treatment regression stage is a critical problem in IV-based methods, underscoring the necessity of decomposing $C$ from observed variables and thereby correcting for confounding bias caused by $C$, coinciding with the findings under the binary scenario. We notice that DRCFR, which performs well on discrete datasets, suffers significantly on the Demand-0-1, reflecting the limitations of their disentanglement theory under the continuous scenario. Among all the baselines, $SD^2$ achieves the best and most stable performance on all Demands Datasets, demonstrating the \textbf{\textit{generalizability}} of $SD^2$ on various types of datasets.

The detailed experimental results on all Demands datasets are provided in the Table \ref{demands}. It can be seen that whether adding the information of instrumental variables (Demands-5-1) or increasing the information of confounders (Demands-0-5), our algorithm performs far better than all the baseline, which proves the \textbf{\textit{efficiency}} and \textbf{\textit{generalizability}} of our method.

\begin{table*}[t]
\vskip 0.15in
\begin{scriptsize}
\begin{center}
    \centering
    \renewcommand\arraystretch{1.6}
    \setlength{\tabcolsep}{4pt}
    \begin{tabular}{c c c c c c c c c}
    \hline
    \textbf{Within-Sample}& \textbf{0-4-4-2-2} & \textbf{2-4-4-2-2} & \textbf{0-4-4-2-10} & \textbf{0-6-2-2-2}&\textbf{Twins-0-16-8}&\textbf{Twins-4-16-8}&\textbf{Twins-0-16-12}&\textbf{Twins-0-20-8} \\ \hline    
    
    \textbf{CLUB}&0.514(0.005)  &0.513(0.006) &0.563(0.006) & 0.481(0.006)&0.020(0.009)&0.029(0.008) & 0.024(0.007)& 0.023(0.005) \\ \hline 
    \textbf{Ours}&\textbf{0.010(0.008)}&\textbf{0.017(0.013)}&\textbf{0.029(0.019)}&\textbf{0.014(0.013)}&\textbf{0.008(0.006)}&\textbf{0.007(0.004)}&\textbf{0.007(0.006)}&\textbf{0.008(0.006)}\\ \hline 
    \textbf{Out-of-Sample}& \textbf{0-4-4-2-2} & \textbf{2-4-4-2-2} & \textbf{0-4-4-2-10} & \textbf{0-6-2-2-2}&\textbf{Twins-0-16-8}&\textbf{Twins-4-16-8}&\textbf{Twins-0-16-12}&\textbf{Twins-0-20-8} \\ \hline    
    
    \textbf{CLUB}&0.517(0.004)  &0.513(0.006) & 0.563(0.004) &0.487(0.005)&0.016(0.013) &0.026(0.009)  &0.021(0.012)  & 0.018(0.010)  \\ \hline \textbf{Ours}& \textbf{0.012(0.008)} & \textbf{0.022(0.017)} & \textbf{0.029(0.018)} & \textbf{0.013(0.010)} &\textbf{0.012(0.007)}&\textbf{0.011(0.006)}&\textbf{0.012(0.006)}&\textbf{0.011(0.009)}\\ \hline
    \hline
    \end{tabular}
\end{center}
\end{scriptsize}
\caption{Performance comparison of bias of ATE between $SD^2$ and  a method which replaces the disentanglement modules in $SD^2$ with one of the SOTA mutual information estimators $CLUB$.}
\label{club}
\end{table*}
\begin{table*}
\begin{scriptsize}
\begin{center}
\renewcommand\arraystretch{1.6}
\setlength{\tabcolsep}{4pt}
    \centering
    \begin{tabular}{ccccccccc}
    \hline
        \multirow{2}{*}{\textbf{Datasets}}   & \multicolumn{4}{c}{\textbf{Within-Sample}}& \multicolumn{4}{c}{\textbf{Out-of-Sample}}  \\  \cline{2-9}   
     & \multicolumn{1}{c}{\textbf{$L_p$}} & \multicolumn{1}{c}{\textbf{$L_p+L_t$}}& \multicolumn{1}{c}{\textbf{$L_p+L_t+L_a$}} & \multicolumn{1}{c}{\textbf{$Total$}} & \multicolumn{1}{c}{\textbf{$L_p$}}& \multicolumn{1}{c}{\textbf{$L_p+L_t$}}& \multicolumn{1}{c}{\textbf{$L_p+L_t+L_a$}}& \multicolumn{1}{c}{\textbf{$Total$}}
           
           \\ \hline
        \textbf{Twins-0-16-8} & 0.027(0.012) & 0.025(0.007) & 0.021(0.007)  & \textbf{0.008(0.006)} & 0.022(0.011) & 0.020(0.011) & 0.018(0.010)  & \textbf{0.012(0.007)}\\  
        \textbf{Twins-4-16-8} & 0.067(0.078) & 0.022(0.005) & 0.024(0.005)  &\textbf{0.008(0.005)} &0.059(0.076) & 0.017(0.012) & 0.019(0.012) & \textbf{0.009(0.006)}\\ 
        \textbf{Twins-0-16-12} & 0.113(0.110) & 0.027(0.005) & 0.024(0.005) & \textbf{0.007(0.006)} & 0.111(0.109) & 0.022(0.011) & 0.019(0.012) & \textbf{0.012(0.006)}\\
        \textbf{Twins-0-20-8} & 0.031(0.010) & 0.020(0.004) & 0.023(0.006) &\textbf{0.008(0.006)}& 0.026(0.013) & 0.015(0.009) & 0.018(0.012) &  \textbf{0.011(0.009)} \\ 
        \textbf{Syn-0-4-4-2-2} & 0.519(0.005) & 0.513(0.005) & 0.513(0.004) & \textbf{0.010(0.008)} & 0.522(0.004) & 0.517(0.004) & 0.516(0.004)  & \textbf{0.012(0.008)}\\ 
        \textbf{Syn-2-4-4-2-2} & 0.517(0.007) & 0.513(0.007) & 0.513(0.007) & \textbf{0.017(0.013)} & 0.517(0.006) & 0.512(0.007) & 0.513(0.006) & \textbf{0.022(0.017)}\\ 
        \textbf{Syn-0-4-4-2-10} & 0.569(0.005) & 0.565(0.004) & 0.565(0.005)  & \textbf{0.029(0.019)} &0.568(0.004) & 0.565(0.004) & 0.564(0.004) & \textbf{0.029(0.018)}\\ 
        \textbf{Syn-0-6-2-2-2} & 0.486(0.004) & 0.483(0.004) & 0.483(0.004)& \textbf{0.015(0.013)} &0.492(0.005) & 0.489(0.005) & 0.489(0.005) & \textbf{0.013(0.010)}\\ \hline
    
    \end{tabular}
\end{center}
\end{scriptsize}
\caption{Ablation Study. $L_p$ represents preserving representation networks and deep outcome classifier only. $L_p+L_t$ adds deep treatment classifier on the basis of the model of $L_p$. $L_p+L_t+L_a$ adds the adjustable variable decomposition module on the basis of the model of $L_p+L_t$. $Total$ model is our proposed model.} 
\label{ablation_study}
\end{table*}

\subsubsection{Disentanglement Visualization}
To check if $VSD^2$ decompose different kinds of underlying factors from pre-treatment variables successfully, similar to \citet{hassanpour2019learning}, we use the first (second) slice to denote the weight matrix that connects the variables in X belonging (not belonging) to the actual variables. The polygons’ radii in Figure \ref{visual} quantify the average weights of the first slice (in red) and the second slice (in blue). We plot the radar charts to visualize the contribution of actual variables to the representations of each factor on $SD^2$ and another typical causal disentangled learning work DRCFR. 

As shown in each sub-figure in Figure \ref{visual}, each vertex on the polygon represents the results of a synthetic dataset $m_v$-$m_z$-$m_c$-$m_a$-$m_u$. Compared with DRCFR, our method realizes much better identification performance of all three underlying factors in all datasets, indicating that our approach can achieve successful disentanglement performance.

\begin{figure*}
\centering
\subfigure[]{
\includegraphics[width=1.6in]{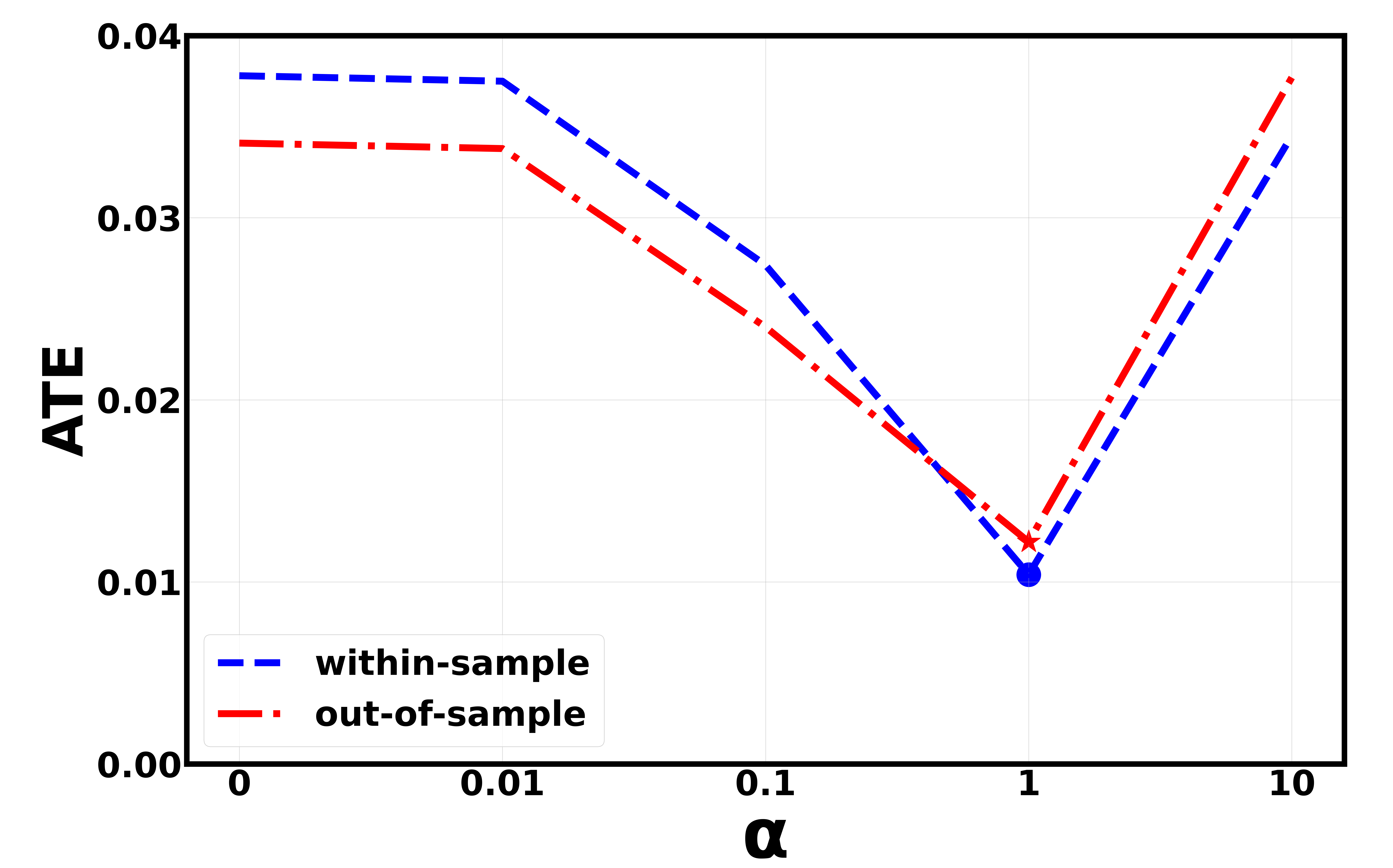}
\label{alpha}
}
\subfigure[]{
\includegraphics[width=1.6in]{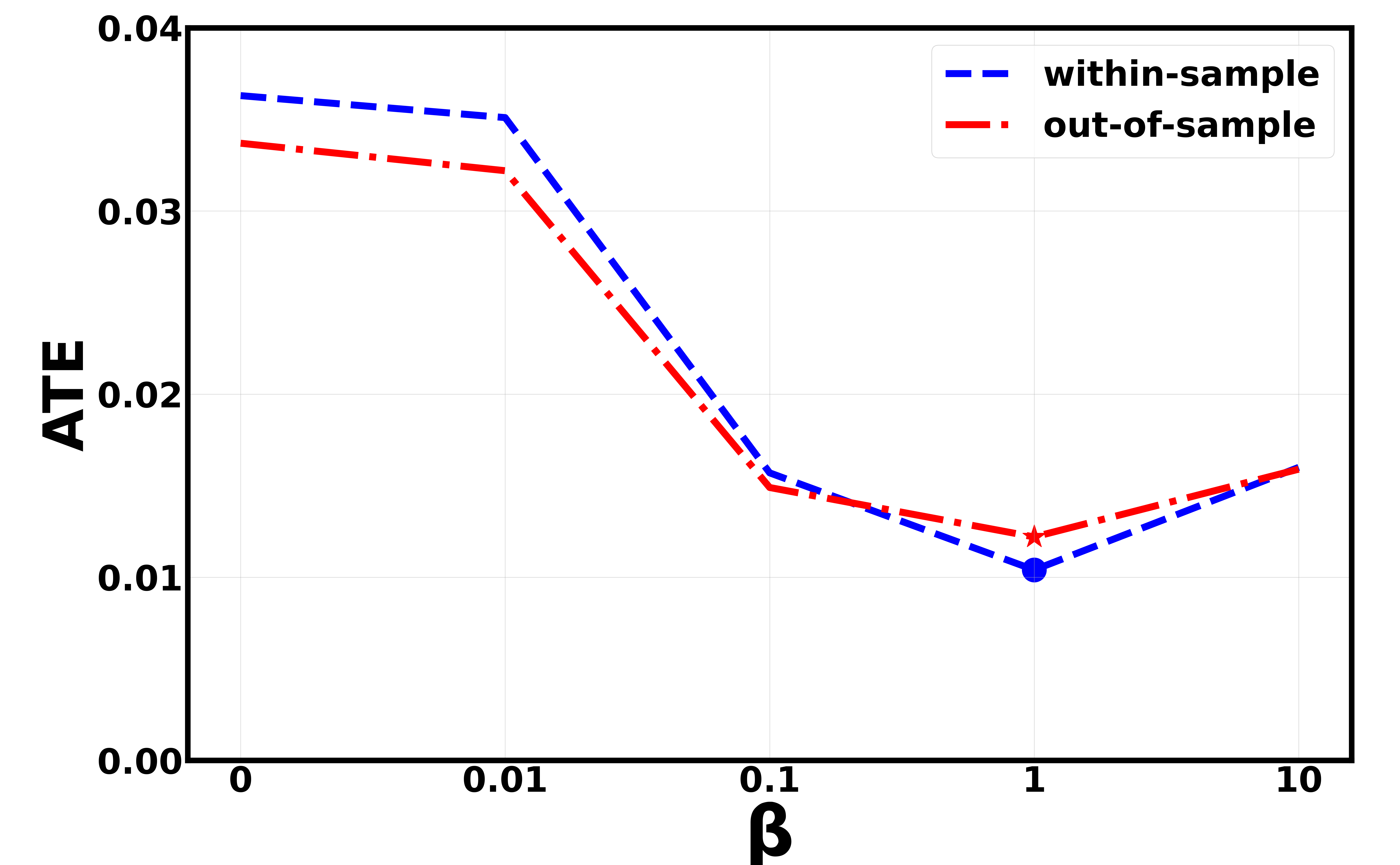}
\label{beta}
}
\subfigure[]{
\includegraphics[width=1.6in]{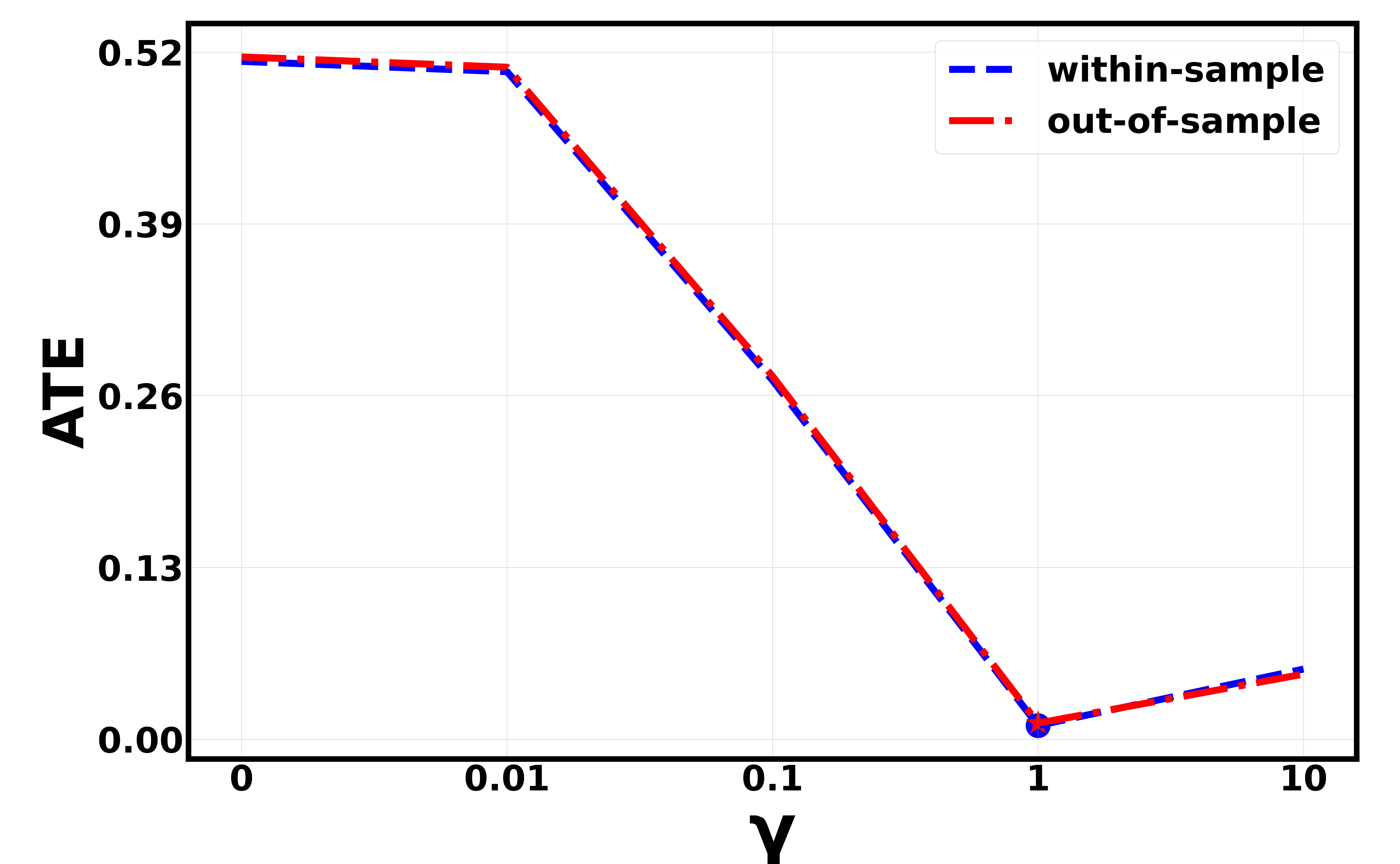}
\label{gamma}
}
\subfigure[]{
\includegraphics[width=1.6in]{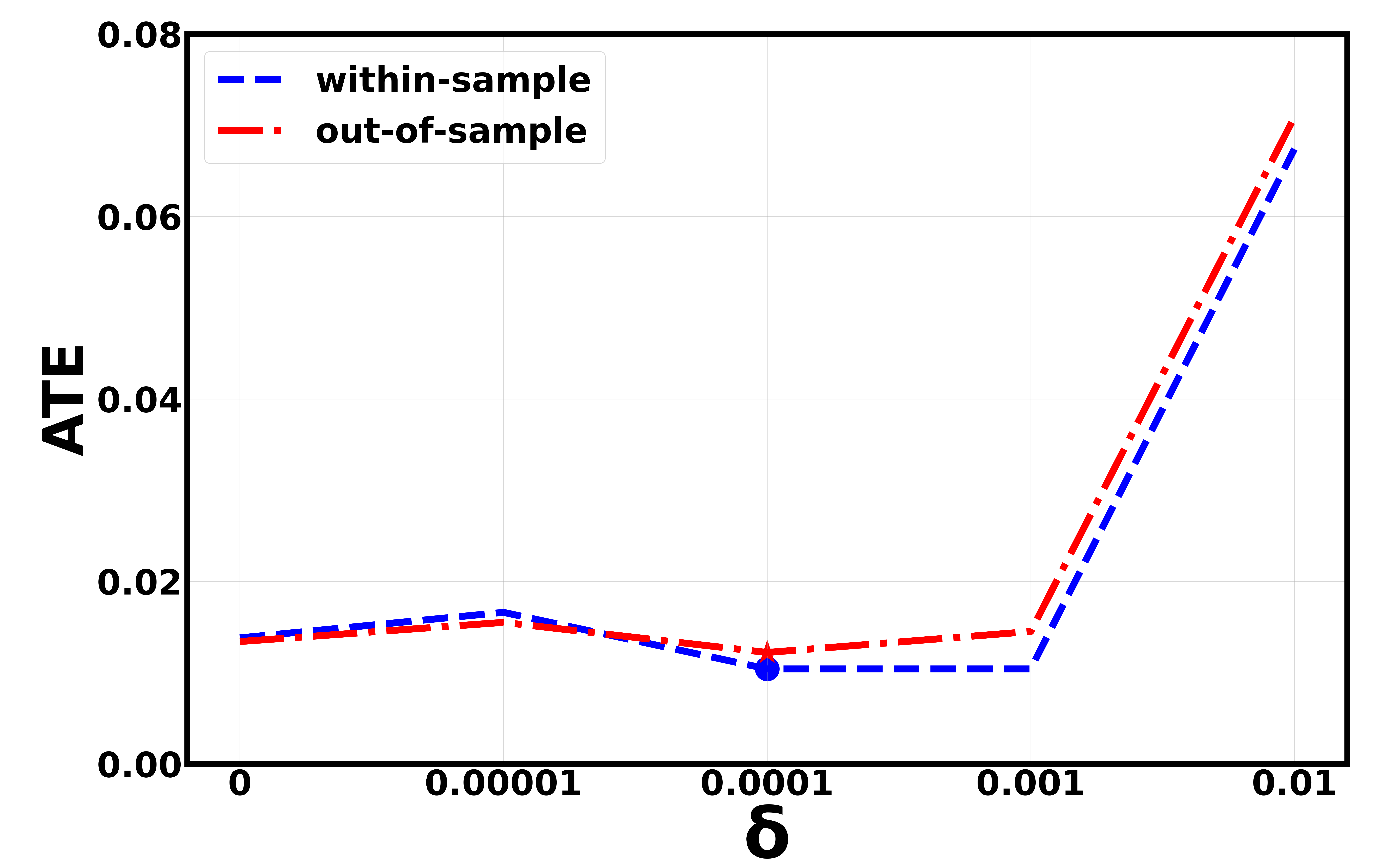}
\label{delta}
}
\caption{Hyper-parameters sensitivity analysis of $\alpha, \beta, \gamma, \delta$ on Syn-0-4-4-2-2 dataset. The blue and red lines show the ATE results of these parameters in within-sample and out-of-sample settings, respectively. The blue circle and red star represent the best parameters for the setting.  }\label{hyper}
\end{figure*}
\begin{figure*}[t]
\centering
\includegraphics[width=0.9\linewidth]{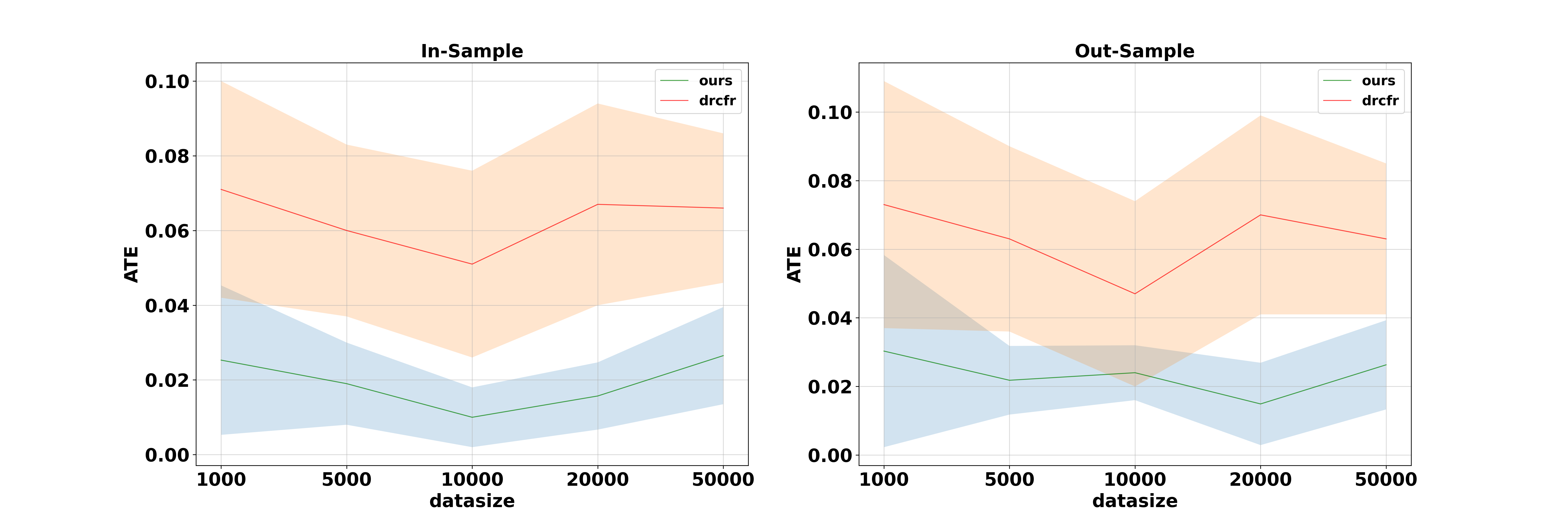}
\caption{Scalability Analysis. The performance of $SD^2$ remains stable and efficient with the variation of sample size in the dataset compared with $DRCFR$, indicating the scalability of $SD^2$. }
\label{scale}
\end{figure*}

\subsubsection{Difficulty of mutual information estimation}
\label{mutual}
To quantify the difficulty of high-dimensional mutual information estimation, we substitute our core disentanglement module with one of the state-of-the-art mutual information estimators, CLUB \citep{Cheng2020CLUBAC}, which has been widely adopted in previous causal disentangled learning methods \citep{wu2020learning, yuan2022auto, cheng2022learning}, aiming to directly minimize the mutual information between $R_z$, $R_c$, and $R_a$. The related experimental results ($CLUB$ in Table \ref{club}) show that the performance of our method gets impeded dramatically, indicating the importance of bypassing the complex mutual information estimation.

\subsubsection{Ablation Study}
\label{abla}
We perform the ablation experiments to examine the contributions of each component in total loss function on final inference performance. The results are shown in Table \ref{ablation_study}. We conduct following steps for the ablation study analysis.

\begin{itemize}
    \item Firstly, we only preserve representation networks and deep outcome prediction networks. The objective loss is reduced to factual loss $L_p$ plus regularization loss.
    \item Secondly, we add the deep treatment prediction networks into the model on the basis of the first model, while the objective loss becomes $L_p+L_t$.
    \item Then, the adjustable variables decomposition module is integrated into the second model, thus the objective loss is $L_p+L_t+L_a$.
    \item Finally, we introduce our shallow treatment and outcome prediction networks into the third model, the objective loss of which is presented in Eq \eqref{loss} in the main paper and named as $Total$.
\end{itemize}

We have the following observations from the experimental results in Table \ref{ablation_study}: 

\begin{itemize}
    \item The model with loss function $L_p$ performs the worst for all data settings on all datasets, demonstrating the significance of treatment prediction for counterfactual prediction in the presence of unobserved confounders.
    \item If we only decompose adjustable variables $A$ from observed features $X$, as the results under the loss $L_p+L_t+L_a$ shows, the model indeed does not get much improvement on the inference performance compared with the results under the loss $L_p+L_t$. This may be due to the fact that only decomposing A cannot eliminate the confounding bias caused by the unmeasured confounders $U$ and underlying confounders $C$ existing in observed features $X$. In addition, it indicates the necessity of the decomposition of $C$ and $Z$ for counterfactual prediction.
    \item When added shallow prediction networks of treatment and outcome during training, as the results of $Total$ model show, our $SD^2$ achieves the best performance with the improvement of more than $60\%$ on Synthetic datasets and over $95\%$ on Twins datasets on the basis of the model under the loss $L_p+L_t+L_a$, which demonstrates the decomposition of $C$ and $Z$ indeed advances the counterfactual inference.
\end{itemize}

\subsubsection{Hyper-Parameters Analysis}

With the multi-term total loss function shown in Eq \eqref{loss}, we study the impact of each item on the counterfactual prediction on Syn-0-4-4-2-2 dataset. As can be seen from Figure \ref{gamma}, the performance of $SD^2$ is mostly affected by the changing in $\gamma$. In addition, although the performance fluctuates with the change of $\alpha$ and $\beta$, $SD^2$ performs better than almost all baselines. These two facts demonstrate that the improvement of inference performance is greatly contributed by the decomposition of $Z$ and $C$. Besides, from Figure \ref{delta}, we find that the performance of the method is affected by changing $\delta$ as well, indicating the necessity of limiting the complexity of the model.

\subsubsection{Scalability Analysis}
\label{scala}

To demonstrate the scalability of our algorithm, we conduct experiments on Syn-0-4-4-2-2 with different sample sizes with our method and DRCFR \citep{hassanpour2019learning}. Figure \ref{scale} shows the results of the related experimental results. It can be seen from the results that the performance of $SD^2$ remains stable and efficient with the variation of sample size in the dataset, which demonstrates the \textbf{\textit{scalability}} of our algorithm.


\section{Conclusion and Limitations}
To resolve the challenge of decomposing mutually independent underlying factors in causal disentangled learning study, we provide a theoretically guaranteed solution to minimizing the mutual information between representations of diverse underlying factors instead of relying on explicit estimation. On this basis, we design a hierarchical self-distilled disentanglement framework $SD^2$ to advance counterfactual prediction by eliminating the confounding bias caused by the observed and unobserved confounders simultaneously. Extensive experimental results on synthetic and real-world datasets validate the \textbf{\textit{effectiveness, generalizability, scalability}} of our proposed theory and framework in both binary and continuous data settings. 

\textbf{Limitations:}  Due to the lack of real-world counterfactual datasets, $SD^2$ is only evaluated on limited tasks. It would be interesting to extend our method into other research areas, such as transfer learning \citep{RojasCarulla2015InvariantMF}, domain adaptation \citep{Magliacane2017DomainAB}, counterfactual fairness \citep{Kim2020CounterfactualFW}, for robust feature extraction. We will leave future work for the generalization of our method to these fields. 
\bibliographystyle{ACM-Reference-Format}
\balance
\bibliography{main}


\begin{thebibliography}{50}


\ifx \showCODEN    \undefined \def \showCODEN     #1{\unskip}     \fi
\ifx \showDOI      \undefined \def \showDOI       #1{#1}\fi
\ifx \showISBNx    \undefined \def \showISBNx     #1{\unskip}     \fi
\ifx \showISBNxiii \undefined \def \showISBNxiii  #1{\unskip}     \fi
\ifx \showISSN     \undefined \def \showISSN      #1{\unskip}     \fi
\ifx \showLCCN     \undefined \def \showLCCN      #1{\unskip}     \fi
\ifx \shownote     \undefined \def \shownote      #1{#1}          \fi
\ifx \showarticletitle \undefined \def \showarticletitle #1{#1}   \fi
\ifx \showURL      \undefined \def \showURL       {\relax}        \fi
\providecommand\bibfield[2]{#2}
\providecommand\bibinfo[2]{#2}
\providecommand\natexlab[1]{#1}
\providecommand\showeprint[2][]{arXiv:#2}

\bibitem[Alaa and van~der Schaar(2017)]%
        {alaa2017bayesian}
\bibfield{author}{\bibinfo{person}{Ahmed~M Alaa} {and} \bibinfo{person}{Mihaela van~der Schaar}.} \bibinfo{year}{2017}\natexlab{}.
\newblock \showarticletitle{Bayesian inference of individualized treatment effects using multi-task gaussian processes}.
\newblock \bibinfo{journal}{\emph{Advances in Neural Information Processing Systems}}  \bibinfo{volume}{30} (\bibinfo{year}{2017}).
\newblock


\bibitem[Almond et~al\mbox{.}(2005)]%
        {almond2005costs}
\bibfield{author}{\bibinfo{person}{Douglas Almond}, \bibinfo{person}{Kenneth~Y Chay}, {and} \bibinfo{person}{David~S Lee}.} \bibinfo{year}{2005}\natexlab{}.
\newblock \showarticletitle{The costs of low birth weight}.
\newblock \bibinfo{journal}{\emph{The Quarterly Journal of Economics}} \bibinfo{volume}{120}, \bibinfo{number}{3} (\bibinfo{year}{2005}), \bibinfo{pages}{1031--1083}.
\newblock


\bibitem[Angrist and Imbens(1994)]%
        {Angrist1994IdentificationAE}
\bibfield{author}{\bibinfo{person}{Joshua~David Angrist} {and} \bibinfo{person}{Guido Imbens}.} \bibinfo{year}{1994}\natexlab{}.
\newblock \showarticletitle{Identification and Estimation of Local Average Treatment Effects}.
\newblock \bibinfo{journal}{\emph{NBER Working Paper Series}} (\bibinfo{year}{1994}).
\newblock


\bibitem[Angrist et~al\mbox{.}(1996)]%
        {angrist1996identification}
\bibfield{author}{\bibinfo{person}{Joshua~D Angrist}, \bibinfo{person}{Guido~W Imbens}, {and} \bibinfo{person}{Donald~B Rubin}.} \bibinfo{year}{1996}\natexlab{}.
\newblock \showarticletitle{Identification of causal effects using instrumental variables}.
\newblock \bibinfo{journal}{\emph{Journal of the American statistical Association}} \bibinfo{volume}{91}, \bibinfo{number}{434} (\bibinfo{year}{1996}), \bibinfo{pages}{444--455}.
\newblock


\bibitem[Belghazi et~al\mbox{.}(2018)]%
        {belghazi2018mutual}
\bibfield{author}{\bibinfo{person}{Mohamed~Ishmael Belghazi}, \bibinfo{person}{Aristide Baratin}, \bibinfo{person}{Sai Rajeshwar}, \bibinfo{person}{Sherjil Ozair}, \bibinfo{person}{Yoshua Bengio}, \bibinfo{person}{Aaron Courville}, {and} \bibinfo{person}{Devon Hjelm}.} \bibinfo{year}{2018}\natexlab{}.
\newblock \showarticletitle{Mutual information neural estimation}. In \bibinfo{booktitle}{\emph{International conference on machine learning}}. PMLR, \bibinfo{pages}{531--540}.
\newblock


\bibitem[Chen et~al\mbox{.}(2016)]%
        {Chen2016InfoGANIR}
\bibfield{author}{\bibinfo{person}{Xi Chen}, \bibinfo{person}{Yan Duan}, \bibinfo{person}{Rein Houthooft}, \bibinfo{person}{John Schulman}, \bibinfo{person}{Ilya Sutskever}, {and} \bibinfo{person}{P. Abbeel}.} \bibinfo{year}{2016}\natexlab{}.
\newblock \showarticletitle{InfoGAN: Interpretable Representation Learning by Information Maximizing Generative Adversarial Nets}. In \bibinfo{booktitle}{\emph{NIPS}}.
\newblock


\bibitem[Cheng et~al\mbox{.}(2022a)]%
        {cheng2022effects}
\bibfield{author}{\bibinfo{person}{Lu Cheng}, \bibinfo{person}{Ruocheng Guo}, \bibinfo{person}{Kasim Candan}, {and} \bibinfo{person}{Huan Liu}.} \bibinfo{year}{2022}\natexlab{a}.
\newblock \showarticletitle{Effects of Multi-Aspect Online Reviews with Unobserved Confounders: Estimation and Implication}. In \bibinfo{booktitle}{\emph{Proceedings of the International AAAI Conference on Web and Social Media}}, Vol.~\bibinfo{volume}{16}. \bibinfo{pages}{67--78}.
\newblock


\bibitem[Cheng et~al\mbox{.}(2022b)]%
        {cheng2022learning}
\bibfield{author}{\bibinfo{person}{Mingyuan Cheng}, \bibinfo{person}{Xinru Liao}, \bibinfo{person}{Quan Liu}, \bibinfo{person}{Bin Ma}, \bibinfo{person}{Jian Xu}, {and} \bibinfo{person}{Bo Zheng}.} \bibinfo{year}{2022}\natexlab{b}.
\newblock \showarticletitle{Learning disentangled representations for counterfactual regression via mutual information minimization}. In \bibinfo{booktitle}{\emph{Proceedings of the 45th International ACM SIGIR Conference on Research and Development in Information Retrieval}}. \bibinfo{pages}{1802--1806}.
\newblock


\bibitem[Cheng et~al\mbox{.}(2020)]%
        {Cheng2020CLUBAC}
\bibfield{author}{\bibinfo{person}{Pengyu Cheng}, \bibinfo{person}{Weituo Hao}, \bibinfo{person}{Shuyang Dai}, \bibinfo{person}{Jiachang Liu}, \bibinfo{person}{Zhe Gan}, {and} \bibinfo{person}{Lawrence Carin}.} \bibinfo{year}{2020}\natexlab{}.
\newblock \showarticletitle{CLUB: A Contrastive Log-ratio Upper Bound of Mutual Information}. In \bibinfo{booktitle}{\emph{International Conference on Machine Learning}}.
\newblock


\bibitem[Chernozhukov et~al\mbox{.}(2013)]%
        {chernozhukov2013inference}
\bibfield{author}{\bibinfo{person}{Victor Chernozhukov}, \bibinfo{person}{Iv{\'a}n Fern{\'a}ndez-Val}, {and} \bibinfo{person}{Blaise Melly}.} \bibinfo{year}{2013}\natexlab{}.
\newblock \showarticletitle{Inference on counterfactual distributions}.
\newblock \bibinfo{journal}{\emph{Econometrica}} \bibinfo{volume}{81}, \bibinfo{number}{6} (\bibinfo{year}{2013}), \bibinfo{pages}{2205--2268}.
\newblock


\bibitem[Chipman et~al\mbox{.}(2010)]%
        {chipman2010bart}
\bibfield{author}{\bibinfo{person}{Hugh~A Chipman}, \bibinfo{person}{Edward~I George}, {and} \bibinfo{person}{Robert~E McCulloch}.} \bibinfo{year}{2010}\natexlab{}.
\newblock \showarticletitle{BART: Bayesian additive regression trees}.
\newblock \bibinfo{journal}{\emph{The Annals of Applied Statistics}} \bibinfo{volume}{4}, \bibinfo{number}{1} (\bibinfo{year}{2010}), \bibinfo{pages}{266--298}.
\newblock


\bibitem[Glass et~al\mbox{.}(2013)]%
        {glass2013causal}
\bibfield{author}{\bibinfo{person}{Thomas~A Glass}, \bibinfo{person}{Steven~N Goodman}, \bibinfo{person}{Miguel~A Hern{\'a}n}, {and} \bibinfo{person}{Jonathan~M Samet}.} \bibinfo{year}{2013}\natexlab{}.
\newblock \showarticletitle{Causal inference in public health}.
\newblock \bibinfo{journal}{\emph{Annual review of public health}}  \bibinfo{volume}{34} (\bibinfo{year}{2013}), \bibinfo{pages}{61}.
\newblock


\bibitem[Goodfellow et~al\mbox{.}(2014)]%
        {Goodfellow2014GenerativeAN}
\bibfield{author}{\bibinfo{person}{Ian~J. Goodfellow}, \bibinfo{person}{Jean Pouget-Abadie}, \bibinfo{person}{Mehdi Mirza}, \bibinfo{person}{Bing Xu}, \bibinfo{person}{David Warde-Farley}, \bibinfo{person}{Sherjil Ozair}, \bibinfo{person}{Aaron~C. Courville}, {and} \bibinfo{person}{Yoshua Bengio}.} \bibinfo{year}{2014}\natexlab{}.
\newblock \showarticletitle{Generative Adversarial Nets}. In \bibinfo{booktitle}{\emph{NIPS}}.
\newblock


\bibitem[Hartford et~al\mbox{.}(2017)]%
        {hartford2017deep}
\bibfield{author}{\bibinfo{person}{Jason Hartford}, \bibinfo{person}{Greg Lewis}, \bibinfo{person}{Kevin Leyton-Brown}, {and} \bibinfo{person}{Matt Taddy}.} \bibinfo{year}{2017}\natexlab{}.
\newblock \showarticletitle{Deep IV: A flexible approach for counterfactual prediction}. In \bibinfo{booktitle}{\emph{International Conference on Machine Learning}}. PMLR, \bibinfo{pages}{1414--1423}.
\newblock


\bibitem[Hassanpour and Greiner(2019)]%
        {hassanpour2019learning}
\bibfield{author}{\bibinfo{person}{Negar Hassanpour} {and} \bibinfo{person}{Russell Greiner}.} \bibinfo{year}{2019}\natexlab{}.
\newblock \showarticletitle{Learning disentangled representations for counterfactual regression}. In \bibinfo{booktitle}{\emph{International Conference on Learning Representations}}.
\newblock


\bibitem[Higgins et~al\mbox{.}(2017)]%
        {higgins2017betavae}
\bibfield{author}{\bibinfo{person}{Irina Higgins}, \bibinfo{person}{Loic Matthey}, \bibinfo{person}{Arka Pal}, \bibinfo{person}{Christopher Burgess}, \bibinfo{person}{Xavier Glorot}, \bibinfo{person}{Matthew Botvinick}, \bibinfo{person}{Shakir Mohamed}, {and} \bibinfo{person}{Alexander Lerchner}.} \bibinfo{year}{2017}\natexlab{}.
\newblock \showarticletitle{beta-{VAE}: Learning Basic Visual Concepts with a Constrained Variational Framework}. In \bibinfo{booktitle}{\emph{International Conference on Learning Representations}}.
\newblock
\urldef\tempurl%
\url{https://openreview.net/forum?id=Sy2fzU9gl}
\showURL{%
\tempurl}


\bibitem[Hoxby(2000)]%
        {hoxby2000does}
\bibfield{author}{\bibinfo{person}{Caroline~M Hoxby}.} \bibinfo{year}{2000}\natexlab{}.
\newblock \showarticletitle{Does competition among public schools benefit students and taxpayers?}
\newblock \bibinfo{journal}{\emph{American Economic Review}} \bibinfo{volume}{90}, \bibinfo{number}{5} (\bibinfo{year}{2000}), \bibinfo{pages}{1209--1238}.
\newblock


\bibitem[Imbens(2004)]%
        {imbens2004nonparametric}
\bibfield{author}{\bibinfo{person}{Guido~W Imbens}.} \bibinfo{year}{2004}\natexlab{}.
\newblock \showarticletitle{Nonparametric estimation of average treatment effects under exogeneity: A review}.
\newblock \bibinfo{journal}{\emph{Review of Economics and statistics}} \bibinfo{volume}{86}, \bibinfo{number}{1} (\bibinfo{year}{2004}), \bibinfo{pages}{4--29}.
\newblock


\bibitem[Kim and Mnih(2018)]%
        {Kim2018DisentanglingBF}
\bibfield{author}{\bibinfo{person}{Hyunjik Kim} {and} \bibinfo{person}{Andriy Mnih}.} \bibinfo{year}{2018}\natexlab{}.
\newblock \showarticletitle{Disentangling by Factorising}. In \bibinfo{booktitle}{\emph{International Conference on Machine Learning}}.
\newblock


\bibitem[Kim et~al\mbox{.}(2020)]%
        {Kim2020CounterfactualFW}
\bibfield{author}{\bibinfo{person}{Hyemi Kim}, \bibinfo{person}{Seungjae Shin}, \bibinfo{person}{Joonho Jang}, \bibinfo{person}{Kyungwoo Song}, \bibinfo{person}{Weonyoung Joo}, \bibinfo{person}{Wanmo Kang}, {and} \bibinfo{person}{Il-Chul Moon}.} \bibinfo{year}{2020}\natexlab{}.
\newblock \showarticletitle{Counterfactual Fairness with Disentangled Causal Effect Variational Autoencoder}.
\newblock \bibinfo{journal}{\emph{ArXiv}}  \bibinfo{volume}{abs/2011.11878} (\bibinfo{year}{2020}).
\newblock


\bibitem[Kingma and Welling(2013)]%
        {Kingma2013AutoEncodingVB}
\bibfield{author}{\bibinfo{person}{Diederik~P. Kingma} {and} \bibinfo{person}{Max Welling}.} \bibinfo{year}{2013}\natexlab{}.
\newblock \showarticletitle{Auto-Encoding Variational Bayes}.
\newblock \bibinfo{journal}{\emph{CoRR}}  \bibinfo{volume}{abs/1312.6114} (\bibinfo{year}{2013}).
\newblock


\bibitem[Li and Yao(2022)]%
        {10027640}
\bibfield{author}{\bibinfo{person}{Xinshu Li} {and} \bibinfo{person}{Lina Yao}.} \bibinfo{year}{2022}\natexlab{}.
\newblock \showarticletitle{Contrastive Individual Treatment Effects Estimation}. In \bibinfo{booktitle}{\emph{2022 IEEE International Conference on Data Mining (ICDM)}}. \bibinfo{pages}{1053--1058}.
\newblock
\urldef\tempurl%
\url{https://doi.org/10.1109/ICDM54844.2022.00130}
\showDOI{\tempurl}


\bibitem[Li and Yao(2024)]%
        {li2024distribution}
\bibfield{author}{\bibinfo{person}{Xinshu Li} {and} \bibinfo{person}{Lina Yao}.} \bibinfo{year}{2024}\natexlab{}.
\newblock \showarticletitle{Distribution-Conditioned Adversarial Variational Autoencoder for Valid Instrumental Variable Generation}. In \bibinfo{booktitle}{\emph{Proceedings of the AAAI Conference on Artificial Intelligence}}, Vol.~\bibinfo{volume}{38}. \bibinfo{pages}{13664--13672}.
\newblock


\bibitem[Lin et~al\mbox{.}(2019)]%
        {lin2019one}
\bibfield{author}{\bibinfo{person}{Adi Lin}, \bibinfo{person}{Jie Lu}, \bibinfo{person}{Junyu Xuan}, \bibinfo{person}{Fujin Zhu}, {and} \bibinfo{person}{Guangquan Zhang}.} \bibinfo{year}{2019}\natexlab{}.
\newblock \showarticletitle{One-stage deep instrumental variable method for causal inference from observational data}. In \bibinfo{booktitle}{\emph{2019 IEEE International Conference on Data Mining (ICDM)}}. IEEE, \bibinfo{pages}{419--428}.
\newblock


\bibitem[Louizos et~al\mbox{.}(2017)]%
        {Louizos2017CausalEI}
\bibfield{author}{\bibinfo{person}{Christos Louizos}, \bibinfo{person}{Uri Shalit}, \bibinfo{person}{Joris~M. Mooij}, \bibinfo{person}{David~A. Sontag}, \bibinfo{person}{Richard~S. Zemel}, {and} \bibinfo{person}{Max Welling}.} \bibinfo{year}{2017}\natexlab{}.
\newblock \showarticletitle{Causal Effect Inference with Deep Latent-Variable Models}. In \bibinfo{booktitle}{\emph{NIPS}}.
\newblock


\bibitem[Magliacane et~al\mbox{.}(2017)]%
        {Magliacane2017DomainAB}
\bibfield{author}{\bibinfo{person}{Sara Magliacane}, \bibinfo{person}{Thijs van Ommen}, \bibinfo{person}{Tom Claassen}, \bibinfo{person}{Stephan Bongers}, \bibinfo{person}{Philip Versteeg}, {and} \bibinfo{person}{Joris~M. Mooij}.} \bibinfo{year}{2017}\natexlab{}.
\newblock \showarticletitle{Domain Adaptation by Using Causal Inference to Predict Invariant Conditional Distributions}. In \bibinfo{booktitle}{\emph{Neural Information Processing Systems}}.
\newblock


\bibitem[Muandet et~al\mbox{.}(2020)]%
        {muandet2020dual}
\bibfield{author}{\bibinfo{person}{Krikamol Muandet}, \bibinfo{person}{Arash Mehrjou}, \bibinfo{person}{Si~Kai Lee}, {and} \bibinfo{person}{Anant Raj}.} \bibinfo{year}{2020}\natexlab{}.
\newblock \showarticletitle{Dual instrumental variable regression}.
\newblock \bibinfo{journal}{\emph{Advances in Neural Information Processing Systems}}  \bibinfo{volume}{33} (\bibinfo{year}{2020}), \bibinfo{pages}{2710--2721}.
\newblock


\bibitem[Pearl et~al\mbox{.}(2000)]%
        {pearl2000models}
\bibfield{author}{\bibinfo{person}{Judea Pearl} {et~al\mbox{.}}} \bibinfo{year}{2000}\natexlab{}.
\newblock \showarticletitle{Models, reasoning and inference}.
\newblock \bibinfo{journal}{\emph{Cambridge, UK: CambridgeUniversityPress}}  \bibinfo{volume}{19} (\bibinfo{year}{2000}), \bibinfo{pages}{2}.
\newblock


\bibitem[Poole et~al\mbox{.}(2019)]%
        {poole2019variational}
\bibfield{author}{\bibinfo{person}{Ben Poole}, \bibinfo{person}{Sherjil Ozair}, \bibinfo{person}{Aaron Van Den~Oord}, \bibinfo{person}{Alex Alemi}, {and} \bibinfo{person}{George Tucker}.} \bibinfo{year}{2019}\natexlab{}.
\newblock \showarticletitle{On variational bounds of mutual information}. In \bibinfo{booktitle}{\emph{International Conference on Machine Learning}}. PMLR, \bibinfo{pages}{5171--5180}.
\newblock


\bibitem[Puli and Ranganath(2020)]%
        {puli2020general}
\bibfield{author}{\bibinfo{person}{Aahlad Puli} {and} \bibinfo{person}{Rajesh Ranganath}.} \bibinfo{year}{2020}\natexlab{}.
\newblock \showarticletitle{General Control Functions for Causal Effect Estimation from IVs}.
\newblock \bibinfo{journal}{\emph{Advances in neural information processing systems}}  \bibinfo{volume}{33} (\bibinfo{year}{2020}), \bibinfo{pages}{8440--8451}.
\newblock


\bibitem[Rojas-Carulla et~al\mbox{.}(2015)]%
        {RojasCarulla2015InvariantMF}
\bibfield{author}{\bibinfo{person}{Mateo Rojas-Carulla}, \bibinfo{person}{Bernhard Sch{\"o}lkopf}, \bibinfo{person}{Richard~E. Turner}, {and} \bibinfo{person}{J. Peters}.} \bibinfo{year}{2015}\natexlab{}.
\newblock \showarticletitle{Invariant Models for Causal Transfer Learning}.
\newblock \bibinfo{journal}{\emph{J. Mach. Learn. Res.}}  \bibinfo{volume}{19} (\bibinfo{year}{2015}), \bibinfo{pages}{36:1--36:34}.
\newblock


\bibitem[Rosenbaum and Rubin(1983)]%
        {rosenbaum1983central}
\bibfield{author}{\bibinfo{person}{Paul~R Rosenbaum} {and} \bibinfo{person}{Donald~B Rubin}.} \bibinfo{year}{1983}\natexlab{}.
\newblock \showarticletitle{The central role of the propensity score in observational studies for causal effects}.
\newblock \bibinfo{journal}{\emph{Biometrika}} \bibinfo{volume}{70}, \bibinfo{number}{1} (\bibinfo{year}{1983}), \bibinfo{pages}{41--55}.
\newblock


\bibitem[Shalit et~al\mbox{.}(2017)]%
        {shalit2017estimating}
\bibfield{author}{\bibinfo{person}{Uri Shalit}, \bibinfo{person}{Fredrik~D Johansson}, {and} \bibinfo{person}{David Sontag}.} \bibinfo{year}{2017}\natexlab{}.
\newblock \showarticletitle{Estimating individual treatment effect: generalization bounds and algorithms}. In \bibinfo{booktitle}{\emph{International Conference on Machine Learning}}. PMLR, \bibinfo{pages}{3076--3085}.
\newblock


\bibitem[Singh et~al\mbox{.}(2019)]%
        {singh2019kernel}
\bibfield{author}{\bibinfo{person}{Rahul Singh}, \bibinfo{person}{Maneesh Sahani}, {and} \bibinfo{person}{Arthur Gretton}.} \bibinfo{year}{2019}\natexlab{}.
\newblock \showarticletitle{Kernel instrumental variable regression}.
\newblock \bibinfo{journal}{\emph{Advances in Neural Information Processing Systems}}  \bibinfo{volume}{32} (\bibinfo{year}{2019}).
\newblock


\bibitem[Veitch et~al\mbox{.}(2020)]%
        {veitch2020adapting}
\bibfield{author}{\bibinfo{person}{Victor Veitch}, \bibinfo{person}{Dhanya Sridhar}, {and} \bibinfo{person}{David Blei}.} \bibinfo{year}{2020}\natexlab{}.
\newblock \showarticletitle{Adapting text embeddings for causal inference}. In \bibinfo{booktitle}{\emph{Conference on Uncertainty in Artificial Intelligence}}. PMLR, \bibinfo{pages}{919--928}.
\newblock


\bibitem[Veitch et~al\mbox{.}(2019)]%
        {veitch2019using}
\bibfield{author}{\bibinfo{person}{Victor Veitch}, \bibinfo{person}{Yixin Wang}, {and} \bibinfo{person}{David Blei}.} \bibinfo{year}{2019}\natexlab{}.
\newblock \showarticletitle{Using embeddings to correct for unobserved confounding in networks}.
\newblock \bibinfo{journal}{\emph{Advances in Neural Information Processing Systems}}  \bibinfo{volume}{32} (\bibinfo{year}{2019}).
\newblock


\bibitem[Wang and Blei(2019)]%
        {wang2019blessings}
\bibfield{author}{\bibinfo{person}{Yixin Wang} {and} \bibinfo{person}{David~M Blei}.} \bibinfo{year}{2019}\natexlab{}.
\newblock \showarticletitle{The blessings of multiple causes}.
\newblock \bibinfo{journal}{\emph{J. Amer. Statist. Assoc.}} \bibinfo{volume}{114}, \bibinfo{number}{528} (\bibinfo{year}{2019}), \bibinfo{pages}{1574--1596}.
\newblock


\bibitem[Wooldridge(2015)]%
        {Wooldridge420}
\bibfield{author}{\bibinfo{person}{Jeffrey~M. Wooldridge}.} \bibinfo{year}{2015}\natexlab{}.
\newblock \showarticletitle{Control Function Methods in Applied Econometrics}.
\newblock \bibinfo{journal}{\emph{Journal of Human Resources}} \bibinfo{volume}{50}, \bibinfo{number}{2} (\bibinfo{year}{2015}), \bibinfo{pages}{420--445}.
\newblock
\showISSN{0022-166X}
\urldef\tempurl%
\url{https://doi.org/10.3368/jhr.50.2.420}
\showDOI{\tempurl}
\showeprint{https://jhr.uwpress.org/content/50/2/420.full.pdf}


\bibitem[Wu et~al\mbox{.}(2022a)]%
        {wu2022instrumental}
\bibfield{author}{\bibinfo{person}{Anpeng Wu}, \bibinfo{person}{Kun Kuang}, \bibinfo{person}{Bo Li}, {and} \bibinfo{person}{Fei Wu}.} \bibinfo{year}{2022}\natexlab{a}.
\newblock \showarticletitle{Instrumental variable regression with confounder balancing}. In \bibinfo{booktitle}{\emph{International Conference on Machine Learning}}. PMLR, \bibinfo{pages}{24056--24075}.
\newblock


\bibitem[Wu et~al\mbox{.}(2022b)]%
        {Wu2022TreatmentEE}
\bibfield{author}{\bibinfo{person}{Anpeng Wu}, \bibinfo{person}{Kun Kuang}, \bibinfo{person}{Ruoxuan Xiong}, \bibinfo{person}{Minqing Zhu}, \bibinfo{person}{Yuxuan Liu}, \bibinfo{person}{Bo Li}, \bibinfo{person}{Furui Liu}, \bibinfo{person}{Zhihua Wang}, {and} \bibinfo{person}{Fei Wu}.} \bibinfo{year}{2022}\natexlab{b}.
\newblock \showarticletitle{Treatment Effect Estimation with Unmeasured Confounders in Data Fusion}.
\newblock \bibinfo{journal}{\emph{ArXiv}}  \bibinfo{volume}{abs/2208.10912} (\bibinfo{year}{2022}).
\newblock


\bibitem[Wu et~al\mbox{.}(2020)]%
        {wu2020learning}
\bibfield{author}{\bibinfo{person}{Anpeng Wu}, \bibinfo{person}{Kun Kuang}, \bibinfo{person}{Junkun Yuan}, \bibinfo{person}{Bo Li}, \bibinfo{person}{Runze Wu}, \bibinfo{person}{Qiang Zhu}, \bibinfo{person}{Yueting Zhuang}, {and} \bibinfo{person}{Fei Wu}.} \bibinfo{year}{2020}\natexlab{}.
\newblock \showarticletitle{Learning decomposed representation for counterfactual inference}.
\newblock \bibinfo{journal}{\emph{arXiv preprint arXiv:2006.07040}} (\bibinfo{year}{2020}).
\newblock


\bibitem[Wu and Fukumizu(2022)]%
        {wu2022betaintactvae}
\bibfield{author}{\bibinfo{person}{Pengzhou~Abel Wu} {and} \bibinfo{person}{Kenji Fukumizu}.} \bibinfo{year}{2022}\natexlab{}.
\newblock \showarticletitle{\${\textbackslash}beta\$-Intact-{VAE}: Identifying and Estimating Causal Effects under Limited Overlap}. In \bibinfo{booktitle}{\emph{International Conference on Learning Representations}}.
\newblock
\urldef\tempurl%
\url{https://openreview.net/forum?id=q7n2RngwOM}
\showURL{%
\tempurl}


\bibitem[Xu et~al\mbox{.}(2020)]%
        {xu2020learning}
\bibfield{author}{\bibinfo{person}{Liyuan Xu}, \bibinfo{person}{Yutian Chen}, \bibinfo{person}{Siddarth Srinivasan}, \bibinfo{person}{Nando de Freitas}, \bibinfo{person}{Arnaud Doucet}, {and} \bibinfo{person}{Arthur Gretton}.} \bibinfo{year}{2020}\natexlab{}.
\newblock \showarticletitle{Learning deep features in instrumental variable regression}.
\newblock \bibinfo{journal}{\emph{arXiv preprint arXiv:2010.07154}} (\bibinfo{year}{2020}).
\newblock


\bibitem[Yao et~al\mbox{.}(2018)]%
        {yao2018representation}
\bibfield{author}{\bibinfo{person}{Liuyi Yao}, \bibinfo{person}{Sheng Li}, \bibinfo{person}{Yaliang Li}, \bibinfo{person}{Mengdi Huai}, \bibinfo{person}{Jing Gao}, {and} \bibinfo{person}{Aidong Zhang}.} \bibinfo{year}{2018}\natexlab{}.
\newblock \showarticletitle{Representation learning for treatment effect estimation from observational data}.
\newblock \bibinfo{journal}{\emph{Advances in Neural Information Processing Systems}}  \bibinfo{volume}{31} (\bibinfo{year}{2018}).
\newblock


\bibitem[Yuan et~al\mbox{.}(2022)]%
        {yuan2022auto}
\bibfield{author}{\bibinfo{person}{Junkun Yuan}, \bibinfo{person}{Anpeng Wu}, \bibinfo{person}{Kun Kuang}, \bibinfo{person}{Bo Li}, \bibinfo{person}{Runze Wu}, \bibinfo{person}{Fei Wu}, {and} \bibinfo{person}{Lanfen Lin}.} \bibinfo{year}{2022}\natexlab{}.
\newblock \showarticletitle{Auto IV: Counterfactual Prediction via Automatic Instrumental Variable Decomposition}.
\newblock \bibinfo{journal}{\emph{ACM Transactions on Knowledge Discovery from Data (TKDD)}} \bibinfo{volume}{16}, \bibinfo{number}{4} (\bibinfo{year}{2022}), \bibinfo{pages}{1--20}.
\newblock


\bibitem[Zhang et~al\mbox{.}(2019)]%
        {zhang2019medical}
\bibfield{author}{\bibinfo{person}{Linying Zhang}, \bibinfo{person}{Yixin Wang}, \bibinfo{person}{Anna Ostropolets}, \bibinfo{person}{Jami~J Mulgrave}, \bibinfo{person}{David~M Blei}, {and} \bibinfo{person}{George Hripcsak}.} \bibinfo{year}{2019}\natexlab{}.
\newblock \showarticletitle{The medical deconfounder: assessing treatment effects with electronic health records}. In \bibinfo{booktitle}{\emph{Machine Learning for Healthcare Conference}}. PMLR, \bibinfo{pages}{490--512}.
\newblock


\bibitem[Zhang et~al\mbox{.}(2021)]%
        {zhang2021treatment}
\bibfield{author}{\bibinfo{person}{Weijia Zhang}, \bibinfo{person}{Lin Liu}, {and} \bibinfo{person}{Jiuyong Li}.} \bibinfo{year}{2021}\natexlab{}.
\newblock \showarticletitle{Treatment effect estimation with disentangled latent factors}. In \bibinfo{booktitle}{\emph{Proceedings of the AAAI Conference on Artificial Intelligence}}, Vol.~\bibinfo{volume}{35}. \bibinfo{pages}{10923--10930}.
\newblock


\bibitem[Zhou et~al\mbox{.}(2023)]%
        {zhou2023emerging}
\bibfield{author}{\bibinfo{person}{Guanglin Zhou}, \bibinfo{person}{Shaoan Xie}, \bibinfo{person}{Guangyuan Hao}, \bibinfo{person}{Shiming Chen}, \bibinfo{person}{Biwei Huang}, \bibinfo{person}{Xiwei Xu}, \bibinfo{person}{Chen Wang}, \bibinfo{person}{Liming Zhu}, \bibinfo{person}{Lina Yao}, {and} \bibinfo{person}{Kun Zhang}.} \bibinfo{year}{2023}\natexlab{}.
\newblock \showarticletitle{Emerging synergies in causality and deep generative models: A survey}.
\newblock \bibinfo{journal}{\emph{arXiv preprint arXiv:2301.12351}} (\bibinfo{year}{2023}).
\newblock


\bibitem[Zhou et~al\mbox{.}(2022)]%
        {zhou2022cycle}
\bibfield{author}{\bibinfo{person}{Guanglin Zhou}, \bibinfo{person}{Lina Yao}, \bibinfo{person}{Xiwei Xu}, \bibinfo{person}{Chen Wang}, {and} \bibinfo{person}{Liming Zhu}.} \bibinfo{year}{2022}\natexlab{}.
\newblock \showarticletitle{Cycle-Balanced Representation Learning For Counterfactual Inference}. In \bibinfo{booktitle}{\emph{Proceedings of the 2022 SIAM International Conference on Data Mining (SDM)}}. SIAM, \bibinfo{pages}{442--450}.
\newblock


\bibitem[Zou et~al\mbox{.}(2020)]%
        {Zou2020JointDA}
\bibfield{author}{\bibinfo{person}{Yang Zou}, \bibinfo{person}{Xiaodong Yang}, \bibinfo{person}{Zhiding Yu}, \bibinfo{person}{B.~V. K.~Vijaya Kumar}, {and} \bibinfo{person}{Jan Kautz}.} \bibinfo{year}{2020}\natexlab{}.
\newblock \showarticletitle{Joint Disentangling and Adaptation for Cross-Domain Person Re-Identification}.
\newblock \bibinfo{journal}{\emph{ArXiv}}  \bibinfo{volume}{abs/2007.10315} (\bibinfo{year}{2020}).
\newblock


\end{thebibliography}
\appendix

\section{Appendix}
\subsection{Theoretical Proof}

\subsubsection{Proof of Theorem \ref{theorem_1}}
\label{4.2}
Consider $R_a$ and $R_c$ as the representations of $A$ and $C$ produced by the representation networks, and let Y be the label of the outcome. We have, 

\begin{equation}
\begin{split}
   &\min I(Y;{R_c}) - I({R_c};Y \mid {R_a}) \Longleftrightarrow \\
   &\min\, H(Y) - H(Y \mid R_c) - H(Y \mid R_a) + H(Y \mid R_c, R_a). 
\end{split}
\label{t1}
\nonumber
\end{equation}
\begin{proof} 
Based on the definition of mutual information \citep{belghazi2018mutual}:
\begin{equation}
I(Y;{R_c}) = H(Y) - H(Y \mid R_c),
\label{t1_1}
\end{equation}
where $H(Y)$ denotes Shannon entropy, and $H(Y \mid R_c)$ is the conditional entropy of $Y$ given $R_c$. Similarly,
\begin{equation}
I({R_c};Y \mid {R_a}) = H(Y \mid R_a) - H(Y \mid R_a, R_c),
\label{t1_2}
\end{equation}
where $I({R_c};Y \mid {R_a})$ denotes the conditional mutual information between $R_c$ and $Y$ given $R_a$; $H(Y \mid R_a)$ and $H(Y \mid R_a, R_c)$ is the the conditional entropy of $Y$ given $R_a$, $R_a$ and $R_c$, respectively.

Combining Eq \eqref{t1_1} and Eq \eqref{t1_2}, we have \textbf{Theo.\ref{theorem_1}} holds.
\end{proof}
\subsubsection{Proof of Corollary \ref{corollary_1}}
\label{4.3}
One of sufficient conditions of minimizing $I({R_a};{R_c})$ is:
\begin{subnumcases}
    {\min}
    {D}_{KL}[\mathcal{P}_{Y}\Vert \mathcal{P}^{R_a}_{Y} ]\nonumber\\
    {D}_{KL}[\mathcal{P}^{R_a}_{Y}\Vert \mathcal{P}^{R_c}_{Y} ],
    \nonumber
\end{subnumcases} 
where $\mathcal{P}^{R_a}_{Y}=p(Y \mid {R_a})$, $\mathcal{P}^{R_c}_{Y}=p(Y\mid{R_c})$ represent the predicted distributions, $\mathcal{P}_{Y}=p(Y)$ represents the real distribution, and $D_{KL}$ denotes the KL-divergence.
\begin{proof}
Based on the definition of conditional entropy, for any continuous variables $R_c$, $R_a$ and $Y$, we have:
\begin{equation}
\begin{split}
      &H(Y) - H(Y \mid R_c) - H(Y \mid R_a) + H(Y \mid R_c, R_a) = \\
      &-\int p(Y)logp(Y) dY \\ 
      &+ \underbrace{\int p(R_c) d{R_c} \int p(Y \mid R_c)logp(Y \mid R_c) dY}_{term C} \\
      &+  \underbrace{\int p(R_a) d{R_a} \int p(Y \mid R_a)logp(Y \mid R_a) dY}_{term A} \\
      &- \underbrace{\iint p(R_a,R_c) d{R_a}d{R_c} \int p(Y \mid {R_a},{R_c})logp(Y \mid {R_a},{R_c}) dY}_{term M}. \\
\end{split}     
\label{integral}
\end{equation}
We further inspect $term C$ in Eq \eqref{integral} and have:
\begin{equation}
\begin{split}
   &\int p(R_c) d{R_c} \int p(Y \mid R_c)logp(Y \mid R_c) dY =\\
   &\iint p(R_c)p(Y \mid R_c)log \left[\frac{p(Y \mid R_c)}{p(Y \mid R_a)}p(Y \mid R_a)\right] d{R_c}dY. \\
\end{split}     
\label{double_int}
\end{equation}
By factorizing the double integrals in Eq \eqref{double_int} into another two components, we show the following:
\begin{equation}
\begin{split}
   &\iint p(R_c)p(Y \mid R_c)log \left[\frac{p(Y \mid R_c)}{p(Y \mid R_a)}p(Y \mid R_a)\right] d{R_c}dY = \\
   &\underbrace{\iint p(R_c)p(Y \mid R_c)log \frac{p(Y \mid R_c)}{p(Y \mid R_a)} d{R_c}dY}_{term C_1} +\\
   & \underbrace{\iint p(R_c)p(Y \mid R_c)log p(Y \mid R_a) d{R_c}dY}_{term C_2}.\\
\end{split}     
\label{factor_c}
\end{equation}
Conduct similar factorization for $term A$ and $term M$ in Eq \eqref{integral}, we have:
\begin{equation}
\begin{split}
    &\int p(R_a) d{R_a} \int p(Y \mid R_a)logp(Y \mid R_a) dY = \\
    &\underbrace{\iint p(R_a)p(Y \mid R_a)log \frac{p(Y \mid R_a)}{p(Y \mid R_c)} d{R_a}dY}_{term A_1} + \\
    &\underbrace{\iint p(R_a)p(Y \mid R_a)log p(Y \mid R_c) d{R_a}dY}_{term A_2}\\
\end{split}
\label{factor_a}
\end{equation}
\begin{equation}
    \begin{split}
    &\iint p({R_a},{R_c}) d{R_a}d{R_c} \int p(Y \mid {R_a},{R_c})logp(Y \mid {R_a},{R_c}) dY = \\
    &\underbrace{\iiint p(R_a,R_c)p(Y \mid {R_a},{R_c})log \frac{p(Y \mid {R_a},{R_c})}{p(Y \mid R_a)} d{R_a}d{R_c}dY}_{term M_1} + \\
    &\underbrace{\iiint p({R_a},{R_c})p(Y \mid {R_a},{R_c})log p(Y \mid R_a) d{R_a}d{R_c}dY}_{term M_2}.\\
\end{split}
\label{factor_m}
\end{equation}
Integrate $term C_1$, $term A_1$ and $term M_1$ over $Y$:
\begin{equation}
    C_1 = \int p(R_c) {D}_{KL}\left[ p(Y \mid R_c) \Vert p(Y \mid R_a) \right] d{R_c},
\label{c_1}
\end{equation}
\begin{equation}
    A_1 = \int p(R_a) {D}_{KL}\left[ p(Y \mid R_a) \Vert p(Y \mid R_c) \right] d{R_a},
\label{a_1}
\end{equation}
\begin{equation}
    M_1 = \iint p({R_a},{R_c}) {D}_{KL}\left[ p(Y \mid {R_a},{R_c}) \Vert p(Y \mid R_a) \right] d{R_a}d{R_c},
\label{m_1}
\end{equation}
where $D_{KL}$ denotes KL-divergence. Integrate $term C_2$ and $term A_2$ over $R_c$ and $R_a$, respectively, we have:
\begin{equation}
    C_2 = \int p(Y)log p(Y \mid R_a) dY,
\label{c_2}
\end{equation}
\begin{equation}
    A_2 = \int p(Y)log p(Y \mid R_c) dY.
\label{a_2}
\end{equation}
Integrate $term M_2$ over $R_c$, we have:
\begin{equation}
    M_2 = \iint p(Y, R_a)log p(Y \mid R_a) d{R_a}dY.
\label{m_2}
\end{equation}
We further factorize Eq \eqref{m_2} into another two components:
\begin{equation}
\begin{split}
    &\iint p(Y, R_a)log p(Y \mid R_a) d{R_a}dY \\
    &= \iint p(Y, R_a)log \left[\frac{p(Y,R_a)}{p(R_a)}\right] d{R_a}dY \\
    &= \iint p(Y, R_a)log p(Y,R_a) d{R_a}dY - \iint p(Y, R_a)log p(R_a) d{R_a}dY\\
    &= -H(Y,{R_a}) + H({R_a})\\
    &= -H(Y\mid{R_a}).
\end{split}
\label{factor_m_2}
\end{equation}
In the view of above, we have the following:
\begin{equation}
\begin{split}
      &H(Y) - H(Y \mid R_c) - H(Y \mid R_a) + H(Y \mid R_c, R_a) = \\
      &-\int p(Y)logp(Y) dY \\ 
      &+ \int p(R_c) {D}_{KL}\left[ p(Y \mid R_c) \Vert p(Y \mid R_a) \right] d{R_c} \\
      &+ \int p(Y)log p(Y \mid R_a) dY - H(Y\mid{R_a}) \\
      &- \iint p({R_a},{R_c}) {D}_{KL}\left[ p(Y \mid {R_a},{R_c}) \Vert p(Y \mid R_a) \right] d{R_a}d{R_c} + H(Y\mid{R_a}).\\
\end{split}     
\label{fill_in}
\end{equation}
Based on the non-negativity of KL-divergence, Eq \eqref{fill_in} is upper bounded by:
\begin{equation}
\begin{split}
      &-\int p(Y)logp(Y) dY + \int p(R_c) {D}_{KL}\left[ p(Y \mid R_c) \Vert p(Y \mid R_a) \right] d{R_c} \\
      &+ \int p(Y)log p(Y \mid R_a) dY=\\
      &\int p(R_c) {D}_{KL}\left[ p(Y \mid R_c) \Vert p(Y \mid R_a) \right] d{R_c}+\\
      &\int p(Y)log \left[\frac{p(Y \mid R_a)}{p(Y)}\right] dY.
\end{split}     
\label{bound}
\end{equation}
Equivalently, we have the upper bound as:
\begin{equation}
\begin{split}
&{\mathbb{E}}_{{R_a}\sim {E_\theta}({R_a}\mid X)}{\mathbb{E}}_{{R_c}\sim {E_\phi}({R_c}\mid X)}[{D}_{KL}[p(Y \mid R_c)\Vert p(Y \mid R_a)]]\\
&+{\mathbb{E}}_{{R_a}\sim {E_\theta}({R_a}\mid X)}\left[log \left[\frac{p(Y \mid R_a)}{p(Y)}\right]\right],
\end{split}
\label{bound_1}
\end{equation}
where $\theta$, $\phi$ denote the parameters of the representation networks of $A$ and $C$, respectively. Therefore, the objective of separating $C$ and $A$ from $X$ can be formalized as:
\begin{equation}
\begin{split}
\min \limits_{\theta,\phi}&{\mathbb{E}}_{{R_a}\sim {E_\theta}({R_a}\mid X)}{\mathbb{E}}_{{R_c}\sim {E_\phi}({R_c}\mid X)}\left[{D}_{KL}[\mathcal{P}^{R_a}_{Y}\Vert \mathcal{P}^{R_c}_{Y}]+log \left[\frac{\mathcal{P}^{R_a}_{Y}}{\mathcal{P}_{Y}}\right]\right],
\end{split}
\label{bound_2}
\end{equation}
where $\mathcal{P}^{R_a}_{Y}=p(Y \mid R_a)$, $\mathcal{P}^{R_c}_{Y}=p(Y \mid R_c)$ and $\mathcal{P}_{Y}$ denote the predicted distributions of $Y$ from the representations $R_a$, $R_c$ and real distribution of $Y$, respectively. 

Clearly, the first term in Eq \eqref{bound_2} is equivalent to minimize the discrepancy between the predicted distributions of $Y$ from the representations $R_a$, $R_c$. Notice the second term in Eq \eqref{bound_2} can be implicitly reduced by minimizing ${D}_{KL}\left[\mathcal{P}^{R_a}_{Y}\Vert \mathcal{P}_{Y}\right]$. Thus, we have:
\begin{subnumcases}
    {\min}
    {D}_{KL}[\mathcal{P}_{Y}\Vert \mathcal{P}^{R_a}_{Y} ] \nonumber\\
    {D}_{KL}[\mathcal{P}^{R_a}_{Y}\Vert \mathcal{P}^{R_c}_{Y} ]   \nonumber 
\end{subnumcases} 

\centerline{$ \Rightarrow \min H(Y) - H(Y \mid R_c) - H(Y \mid R_a) + H(Y \mid R_c, R_a).$}
\textbf{Corol.\ref{corollary_1}} holds.
\end{proof}

\subsection{Reproducibility}

\subsubsection{Loss Functions for Continuous Scenario}
\label{conti}
The loss function for disentangling A is defined as following:
\begin{equation}
    \mathcal{L}_{a} = {L}(\hat{{T}_{c}},{T})+ {KL}(\hat{{T}_{c}},\hat{{T}})+
   {KL}(\hat{{T}_{c}},\hat{{T}_{a}}),     
\label{L_a}
\end{equation}
where $\hat {T}_{variable}$ in $L(\cdot)$ represents the predicted values for $variable$ while in $KL(\cdot)$ it denotes the distribution of the variable. Same below.

We assume that the continuous variable follows a normal distribution, therefore the KL divergence can be calculated with:
\begin{equation}
\mathrm{KL}(q \| p)=\log \sigma_2-\log \sigma_1+\frac{\sigma_1^2+\left(\mu_1-\mu_2\right)^2}{2 \sigma_2^2}-\frac{1}{2},
\label{kl}
\end{equation}
where $\mathbf{q} \sim \mathcal{N}\left(\mu_1, \sigma_1^2\right), \mathbf{p}\sim\mathcal{N}\left(\mu_2, \sigma_2^2\right)$.

To reduce the confounding bias led by observed confounders, we define the following loss function:
\begin{equation}
    \mathcal{L}_{oc} = L(\hat{{T}_{z}},{T})+ 
    KL(\hat{{T}_{z}},\hat{{T}})+
    KL(\hat{{T}_{z}},\hat{{T}_{\tilde {c}}}),   
\label{L_oc}
\end{equation}
where $\tilde c$ denotes the representation of $C$ after re-balance network.

Therefore, the total loss function of $SD^2$ for the continuous scenario can be devised as:
\begin{equation}
\begin{split}
\mathcal{L}_{SD^2}= &\underbrace{ L(\hat{Y},Y)+\alpha L(\hat{T},T)}_{factual\; loss}+ 
\underbrace{\beta \mathcal{L}_{a} + \gamma (\mathcal{L}_{c} +  \mathcal{L}_{z})}_{disentanglement\; loss}\\
&+\underbrace{\omega \mathcal{L}_{oc}}_{rebalance\; loss}+
\underbrace{\delta {{\parallel}W{\parallel}}_{2}}_{regularization\; loss}.
\end{split}
\label{loss_conti}
\end{equation}

\subsubsection{Hardware}
\label{hardware}
In this work, we perform all experiments on a cluster with two 12-core Intel Xeon E5-2697 v2 CPUs and a total 768 GiB Memory RAM.

\end{document}